\documentclass[%
aip,
amsmath,amssymb,
reprint,%
author-year,%
]{revtex4-1}

\usepackage{graphicx}
\usepackage{dcolumn}
\usepackage{bm}
\usepackage{multirow}
\renewcommand{\arraystretch}{1.2}
\usepackage{url}
\usepackage[utf8]{inputenc}
\usepackage[T1]{fontenc}
\usepackage{mathptmx}
\usepackage{etoolbox}
\usepackage[hidelinks]{hyperref}

\makeatletter
\def\@email#1#2{%
	\endgroup
	\patchcmd{\titleblock@produce}
	{\frontmatter@RRAPformat}
	{\frontmatter@RRAPformat{\produce@RRAP{*#1\href{mailto:#2}{#2}}}\frontmatter@RRAPformat}
	{}{}
}%
\makeatother

\usepackage{amsthm}
\usepackage{bbold}
\usepackage{flushend}
\usepackage{siunitx}
\usepackage{yhmath} 

\usepackage{xcolor}
\definecolor{sina}{rgb}{0.0,0.0,1.0} 

\definecolor{eva}{rgb}{0.0,0.0,0.0} 
\def\eva{\textcolor{eva}}

\definecolor{chris}{rgb}{0.0,0.0,0.0} 
\def\chris{\textcolor{chris}}

\definecolor{kathrin}{rgb}{0.3,0.0,0.5} 

\definecolor{comment}{rgb}{0.2,0.5,0.3}

\def\p{\partial }
\def\R{\mathbb{R}}
\def\D{\mathrm{D}}
\def\d{\mathrm{d}}
\theoremstyle{definition}
\newtheorem{exmp}{Example}[section]
\newtheorem{rem}{Remark}[section]
\begin{document}
\preprint{AIP/123-QED}

\title[Hamiltonian Neural Networks with Automatic Symmetry Detection]
{Hamiltonian Neural Networks with Automatic Symmetry Detection} 

\author{Eva Dierkes} \email{eva.dierkes@uni-bremen.de}
\affiliation{Center for Industrial Mathematics, University of Bremen, Germany, }
\author{Christian Offen} 
\author{Sina Ober-Blöbaum}
\affiliation{Department of Mathematics, Paderborn University, Germany}
\author{Kathrin Flaßkamp}
\affiliation{Systems Modeling and Simulation,
	Saarland University, Germany}
	
%
%
%

\begin{abstract}                
Recently, Hamiltonian neural networks (HNN) have been introduced to incorporate prior physical knowledge when learning the dynamical equations of Hamiltonian systems. %
Hereby, the symplectic system structure is preserved despite the data-driven modeling approach. %
However, preserving symmetries 
requires additional attention. %
In this research, we enhance HNN with a Lie algebra framework to detect and embed symmetries in the neural network. %
This approach allows to simultaneously learn the symmetry group action and the total energy of the system.
As illustrating examples, a pendulum on a cart and a two-body problem from astrodynamics are considered.
\end{abstract}


\maketitle
\begin{quotation}
Incorporating physical system knowledge into data-driven system identification has been shown to be beneficial. %
The approach presented in this article combines learning of an energy-conserving model from data with detecting a Lie group representation of the unknown system symmetry. %
The proposed approach can improve the learned model and reveal underlying symmetry simultaneously. %
\end{quotation}
\section{Introduction}

\chris{Modeling mechanical systems from first principles as Hamiltonian or Lagrangian systems or using a Newton-Euler modeling approach has a long history.
Recently, data-driven techniques have gained attention within this context to describe complex physical systems for which either no model exists or existing models are too complicated to use in simulations. Examples include novel materials or the identification of suitable reduced-order models for} multi-physics systems.

Mechanical systems are well-known to possess characteristic properties, such as symplecticity of the Hamiltonian/Lagrangian flow, symmetries which lead to the preservation of momentum maps, and energy-preservation in the absence of external forcing. %
In physics-based modeling, \chris{models are derived from first principles. For numerical computations they are discretized using} structure-preserving \chris{techniques such as symplectic or } variational integrators. \chris{This makes sure that the discrete model shares structural properties with the original, continuous system }\citep[see][]{marsden2001discrete}. %

In data-based modeling, standard methods such as neural network approximations of vector fields or flows \chris{typically violate structural constraints of the true system such as symmetries or conservation laws.} %
The SINDy approach, proposed by \cite{brunton_discovering_2016} applies sparse regression to get a symbolic representation of the system's ODE based on suitable basis functions, without focusing on mechanical systems. %
\cite{udrescu_ai_2020} presented an approach where deep learning is used to find symmetries in the data in order to reduce the exponentially large search space of all possible basis functions for a symbolic system representation. %

Learning of Hamiltonian systems is addressed \eva{among others in \cite{bertalan19On,greydanus_hamiltonian_2019,Zhong2020Symplectic}. %
In \cite{Zhong2020Symplectic}, }the scalar valued Hamiltonian is learned by using neural networks to model the typical components of the Hamiltonian separately, such as the potential energy or the mass matrix. %
A more general approach was proposed by \cite{greydanus_hamiltonian_2019} and named \emph{Hamiltonian Neural Networks (HNN)}. \chris{When} the Hamiltonian is learned, the system's symplecticity and energy conservation are preserved by design. %
\eva{\cite{bertalan19On} shows a framework in which symplectic coordinates and a Hamiltonian are learned simultaneously. They also treat high dimensional data, in particular videos of mechanical systems.}

Extending the HNN approach, \cite{dierkes2021learning} show how to learn a symmetry-preserving Hamiltonian, if the system symmetry is known a priori. %
\eva{\cite{finzi2020generalizing} showed how neural networks can be made equivariant and symmetric utilising convolutional layers with symmetric kernels. }

Rather than learning the continuous Hamiltonian, \cite{ChenZAB20,SSI} learn a Hamiltonian tailored to symplectic integration schemes. In this way, discretization errors in the integration step are eliminated and trajectory observations instead of observations of the Hamiltonian vector field can be used for training. A similar strategy can be employed on the Lagrangian side \citep[see][]{LSI, LDensityLearning}. 
\eva{An \chris{approach based on symplectic Lie group integrators on $T^\ast \mathrm{SE}^{\otimes N}$} can be found in \cite{So22datadriven}. }%

Another approach to preserve symplectic structure when learning dynamical systems is to learn the system's Hamiltonian flow map by learning its generating function \citep[e.g.,][]{chen21datadriven,Rath2021} or by using symplectic neural networks \citep[e.g.,][]{SympNets}, where symplecticity is guaranteed by the network architecture.
\chris{In the context of Gaussian Processes, group integration matrix kernels (GIM kernels) can be used to enforce symmetry properties of learned quantities as demonstrated, for instance, in }\cite{Ridderbusch2021}.
Lie algebra convolutional neural networks (CNN) by \citep{Dehmany2021} can automatically discover symmetries and preserve them by using suitable localizations of the kernels present in CNN and making them group invariant. The approach SymDLNN \citep[cf.][]{lishkova_discrete_2022} discovers continuous symmetries and conservation laws of Euler-Lagrange dynamics while learning a symmetric time-discrete Lagrangian from snapshot data of motions. 

In this article, we will learn the Hamiltonian of a system and use a Lie algebra framework to detect and embed symmetries into a neural network that models the Hamiltonian of a dynamical system.
For Lie group actions, such as the action by affine linear transformations \chris{on the configuration space}, our network automatically detects subgroups under which the Hamiltonian function is invariant. This is achieved by learning the Hamiltonian along with a spanning set of generators of invariant vector fields and testing via a loss function whether the derivatives of the Hamiltonian $H$ along the invariant vector fields vanish.
Utilising Noether's theorem, the two ingredients, symmetries and symplectic structure, then allow us to identify integrals of motions. %
We exemplify the concept by the cart-pendulum and the two-body problem, for which we train a symmetry-symplecticity-preserving neural network.

\section{Preliminaries}
To introduce notation and our main examples, we summarize preliminaries from the referred research fields, namely geometric mechanics with symmetries and neural network \chris{machine} learning \chris{as needed in the context of this article}.
We refer to \cite{Marsden78,MarsdenRatiu,Olver1986, Goodfellow-et-al-2016} for details \chris{and generalisations}.

\subsection{Symplectic maps and Hamiltonian dynamics}

Let $Q$ denote an open, non-empty subset of $\R^{n}$, which we refer to as {\em configuration space}.
The cotangent bundle $\mathcal{M} = T^\ast Q$ can be identified with an open subset of $\R^{2n} \cong \R^n \times \R^n$ with (Darboux) coordinates $z=(q,p) \in \mathcal{M}$. The manifold $\mathcal{M}$ is called {\em phase space}.
A diffeomorphism $\Psi \colon \mathcal{M} \to \mathcal{M}$ is {\em symplectic}, if $\D \Psi(z)^\top J \D \Psi(z)=J$ for all $z \in \mathcal{M}$, where $\D \Psi(z)$ is the Jacobi matrix of $\Psi$ at $z$ and
\[
J = \begin{pmatrix}
0&-I_n\\
I_n & 0
\end{pmatrix} \in \R^{2n}
\]
is the symplectic structure matrix of $\mathcal{M}$. Here, $I_n$ denotes the $n$-dimensional identity matrix.

Symplectic maps on $\mathcal{M}$ can be constructed as cotangent lifts of diffeomorphisms and as flows of Hamiltonian vector fields:
given a diffeomorphism $\phi \colon Q \to Q$ we can obtain a symplectic map $\Psi \colon \mathcal{M} \to \mathcal{M}$ as $\Psi(q,p)=(\phi^{-1}(q),\D \phi(q)^\top p)$. The map $\Psi \colon \mathcal{M} \to \mathcal{M}$ is called {\em the cotangent lift of $\phi$}.

For any scalar valued map $H \colon \mathcal{M} \to \R$ we consider the {\em Hamiltonian vector field} $X_H(z) = J^{-1} \nabla H(z)$. {\em Hamilton's equations} are given as the flow equations to $X_H$, i.e.\
\begin{equation}\label{eq:HamEQ}
\dot{z} = X_H(z) = J^{-1}\nabla H(z)
\quad
\text{or}
\quad
\begin{cases}
\dot{q} &= \phantom{-}\nabla_p H(q,p)\\
\dot{p} &= -\nabla_q H(q,p)
\end{cases}.
\end{equation}
For $\tau \in \R$, the flow map $\Psi_\tau$
assigns to $z_0 \in \mathcal{M}$ the solution of \eqref{eq:HamEQ} at time $\tau$ subject to the initial condition $z(0)=z_0$ (if it exists). On its domain of definition $\mathcal{M}_\tau \subset \mathcal{M}$, the map $\Psi_\tau \colon \mathcal{M}_\tau \to \Psi_\tau(\mathcal{M})$ is symplectic.
Symplectic maps on $\mathcal{M}$ form a group under composition which we denote by $\mathrm{Symp}(\mathcal{M})$.

\subsection{Lie group actions}\label{ssec:LieGroup}
Let us briefly introduce Lie group actions and invariant vector fields. For details we refer to \cite{MarsdenRatiu}.

Lie groups are algebraic groups which are also smooth manifolds.
%
%
For a Lie group $G$, the Lie algebra $\mathfrak{g}$ is the tangent space of $G$ at the neutral element $e$, i.e., for
\begin{align*}
\mathfrak{g} = \Big\{v= \left.\left.\frac{\d}{\d t}\right|_{t=0} \gamma(t) \, \right|  & \, \gamma \colon (-\epsilon,\epsilon) \to G \text{ smooth}\\
&  \text{with } \gamma(0)=e, \, \epsilon>0 \Big\}.
\end{align*}
Denote by $\exp \colon \mathfrak{g} \to G$ the {\em exponential map}: in case $G$ is a subgroup of the group of invertible matrices $\mathrm{GL}(\R,n) = \{ A \in \R^{n \times n}| \det(A) \not = 0 \}$, the map $\exp$ is given by the matrix exponential.

A {\em Lie group action} on a manifold $\mathcal M$ is a group homomorphism $L \colon G \to \mathrm{Diff}(\mathcal{M}), \, g \mapsto L_g$, i.e.\ for all $g,h\in G$ the map $L_g \colon \mathcal{M} \to \mathcal{M}$ is a diffeomorphism and $L_{g \circ h} = L_g \circ L_h$. If the manifold $\mathcal M$ is symplectic and each $L_g$ is a symplectic map, then the Lie group action is called {\em symplectic}.


To each Lie algebra element $v \in \mathfrak{g}$ we associate the {\em fundamental vector field} $\widehat{v}$ defined by
\[
\widehat{v}_z = \left.\frac{\d}{\d t}\right|_{t=0} L_{\exp{(tv)}}(z) \in T_z \mathcal{M}, \quad z \in \mathcal{M}.\]
These vector fields can be thought of as infinitesimal actions of the Lie group $G$ on $\mathcal{M}$.

Of central importance for the applications considered in this article are actions of subgroups of the group of affine linear transformations, which are introduced in the following.

\begin{exmp}[Affine linear transformations]\label{ex:affineLinear}
Let $Q = \R^n$ and $\mathcal{M} = T^\ast Q \cong \R^{2n}$ with Darboux coordinates $(q,p)$. The group of affine transformations $G=\mathrm{Aff}(Q)$ on $Q$ can be represented as
\[
G =\mathrm{Aff}(Q)= \left\{ \left.\begin{pmatrix}
A&b\\ 0 & 1
\end{pmatrix} \, \right| \, A \in \mathrm{Gl}(\R,n), \, b \in \R^n \right\},
\]
where the group operation is matrix multiplication and $\mathrm{Gl}(\R,n)$ denotes the group of invertible matrices. Its Lie algebra is 
\[
\mathfrak{g} =\mathrm{aff}(Q) = \left\{ \left.\begin{pmatrix}
	M&b\\ 0 & 0
\end{pmatrix} \, \right| \, M \in \R^{n \times n}, \, b \in \R^n \right\}.
\]
As a shorthand for $\begin{pmatrix}	A&b\\ 0 & 1\end{pmatrix} \in G$ and for $\begin{pmatrix}	M&b\\ 0 & 0\end{pmatrix} \in \mathfrak{g}$ we may write $(A,b) \in G$ or $(M,b) \in \mathfrak{g}$, respectively. 
\chris{On $\mathfrak{g} = \mathrm{aff}(Q)$ we can consider the inner product}
\begin{equation}\label{eq:FrobNormAff}
\begin{split}
\langle v^{(1)},v^{(2)}\rangle_{\mathrm{aff}(Q)}
&=
\langle (M^{(1)},b^{(1)}),(M^{(2)},b^{(2)}) \rangle_{\mathrm{aff}(Q)}\\
&= \langle M^{(1)},M^{(2)}\rangle + \langle b^{(1)},b^{(2)} \rangle
\end{split}
\end{equation}
for $v^{(j)} = (M^{(j)},b^{(j)}) \in \mathfrak g$. Here $\langle M^{(1)},M^{(2)}\rangle$ denotes the Frobenius inner product $\langle M^{(1)},M^{(2)}\rangle = \sum_{i,j=1}^n M^{(1)}_{ij}M^{(2)}_{ij}$ on $\R^{n \times n}$ and $\langle b^{(1)},b^{(2)} \rangle$ the Eucledian inner product on $\R^n$. This definition of $\langle v^{(1)},v^{(2)}\rangle_{\mathrm{aff}(Q)}$ corresponds to the Frobenius inner product of the matrices $\begin{pmatrix}	M^{(1)}&b^{(1)}\\ 0 & 0\end{pmatrix},\begin{pmatrix}	M^{(2)}&b^{(2)}\\ 0 & 0\end{pmatrix} \in \R^{(n+1) \times (n+1)}$. The induced norm on $\mathfrak{g}$ is given as $\| v\|_{\mathrm{aff}(Q)} = \langle v,v\rangle_{\mathrm{aff}(Q)}$ for $v \in \mathfrak g$

The exponential map $\exp \colon \mathfrak{g} \to G$ is given as the matrix exponential. The group $G$ acts on $Q$ by affine linear transformations $(A,b) \mapsto \ell_{(A,b)}$ with $\ell_{(A,b)}(q) = Aq+b$. Furthermore, the group $G$ acts on $\mathcal{M}=T^\ast Q$ symplectically with $(A,b) \mapsto L_{(A,b)}$ where $L_{(A,b)} \colon \mathcal{M} \to \mathcal{M}$ is the contangent lift of $\ell_{(A,b)} \colon Q \to Q$, i.e.\ 
\[
L_{(A,b)}(q,p) = \begin{pmatrix}
A^{-1}(q-b),& A^\top p
\end{pmatrix}.
\]
For $v = (M,b) \in \mathfrak{g}$ the invariant vector field $\widehat v = \widehat{(M,b)}$ 
at $(q,p) \in \mathcal{M}$ is given as
\[
\widehat v_{(q,p)}
=
\widehat{(M,b)}_{(q,p)} = \begin{pmatrix}
-Mq-b, & M^\top p
\end{pmatrix}.
\]
Vector fields can be interpreted as derivations: if $\widehat{v}=\widehat{(M,b)}_{(q,p)}$ is applied to $H \colon \mathcal{M} \to \R$ we obtain the directional derivative
\begin{equation}\label{eq:AffTraforCheck}
	\begin{split}
	\widehat{v}_{(q,p)}(H)
	&=\widehat{(M,b)}_{(q,p)}(H)\\
	&=(-Mq-b)^\top \nabla_q H(q,p) + (M^\top p)^\top \nabla_p H(q,p).
	\end{split}
\end{equation}
\end{exmp}

\subsection{Symmetry actions and conserved quantities}\label{sec:SymmetriesConservedQuantities}

Consider a Hamiltonian $H \colon \mathcal{M} \to \R$ on the symplectic manifold $\mathcal{M} = T^\ast Q$, where $Q$ is an open, non-empty subset of $\R^n$. 
A symplectic Lie group action $L \colon G \to \mathrm{Symp}(\mathcal{M})$ is a {\em symmetry} if $H \circ L_g = H$ for all $g \in G$. This has important consequences for the dynamical system defined by the Hamiltonian vector field $X_H$: if $\gamma \colon [a,b] \to \mathcal{M}$ is a motion, i.e.\ $\gamma$ fulfills Hamilton's equations \eqref{eq:HamEQ}, then $L_g \circ \gamma$ is a motion as well. 
Moreover, if $v \in \mathfrak{g}$ such that $\widehat{v}$ is the Hamiltonian vector field to a smooth map $I \colon \mathcal{M} \to \R$ then $I$ is a conserved quantity of the system, i.e.\ $I \circ \gamma = I$ for all motions $\gamma$ (Noether's theorem).

\begin{exmp}\label{ex:AffLinearConserved}
If a Hamiltonian $H$ is invariant under a translation in the direction of $b \in \R^n$, i.e.\ the directional derivative $\nabla_{(b,0)} H(q,p) = \sum_{j=1}^n b^j \frac{\p H}{ \p q^j}(q,p)$ vanishes for all $(q,p) \in \mathcal{M}$, then the quantity $I(q,p)= b^\top p$ is conserved under motions. More generally, if $(M,b) \in \mathfrak{g}$ is an element of the Lie algebra of affine linear transformations and 
$\widehat{(M,b)}_{(q,p)}(H) =0$ for all $(q,p) \in \mathcal{M}$ (see \eqref{eq:AffTraforCheck}) then
\begin{equation}\label{eq:ConservedAff}
I(q,p)=p^\top(-Mq-b)   
\end{equation}
is a conserved quantity.
\end{exmp}

\begin{rem}\label{rem:BracketClosure}
If for a Lie group action on a Hamiltonian system with Hamiltonian $H$ the fundamental vector fields $\widehat{v}^1, \widehat{v}^2$ to Lie algebra elements $v^1,v^2 \in \mathfrak{g}$ generate infinitesimal symmetries, i.e.\ $\widehat{v}^1(H)=0$ and $\widehat{v}^2(H)=0$, then, the commutator of the vector fields $[\widehat v^1, \widehat v^2]$ is the fundamental vector field to a Lie algebra element $[v^1,v^2] \in \mathfrak{g}$ and $\widehat{[v^1,v^2]}(H)=0$.
\end{rem}

The following Examples~\ref{ex:PendCart} and \ref{ex:TwoBody} introduce the two main examples to which we will apply our framework.

\begin{exmp}[Pendulum on a cart]\label{ex:PendCart}
The proposed framework will be applied to the example of a planar pendulum mounted on a cart  \citep[cf.\ e.g.][]{Bloch}. %
The generalized coordinates are $q=(s,\varphi)$, where $s$ is the position of the cart and $\varphi$ \chris{denotes} the angle \chris{of the pendulum with the upright} vertical. %
Since this system is well studied, we can use the true Hamiltonian for, firstly, generating data points and, secondly, evaluating the performance of our data-driven approach. %
The Hamiltonian is given by
\begin{align}\label{eq:Hamiltonian_PendCart}
    H(\varphi,p_s,p_\varphi) = \frac{ap_s^2+ 2bp_sp_\varphi\cos\varphi + c p_\varphi^2}{2ac-b^2\cos^2\varphi} - D\cos\varphi,
\end{align}
with constants $a=ml^2$, $b=ml$, $c=m_0+m$ and $D=-mgl$. Here $l$ is the length and $m$ the mass of the pendulum, $m_0$ corresponds to the mass of the cart and $g=9.81$ is the gravitation constant. In our numerical experiments $m=m_0=l=1$.

The Hamiltonian $H$ does not explicitly depend on the cart's position $s$. %
Therefore, $H$ is invariant under translations in $s$, i.e.\ the symplectic transformations $(s,\phi,p_s,p_\phi) \mapsto (s+b_1,\phi,p_s,p_\phi)$ for all $b_1 \in \R$. %
This can be interpreted as actions by the following subgroup of $\mathrm{Aff}(Q)$ (see Example~\ref{ex:affineLinear}):
\[
G=\left\{(I_2,b)=\left. \begin{pmatrix}I_2 & b\\ 0&1 \end{pmatrix} \right| b= \begin{pmatrix} b_1\\0\end{pmatrix}, b_1 \in \R \right\} \subset \mathrm{Aff}(Q).
\]
Here $I_2$ is the 2-dimensional identity matrix.
Its Lie algebra $\mathfrak{g}$ is given as the following sub-Lie algebra of $\mathrm{aff}(Q)$
\begin{align}\label{eq:LieAlgebra_PendCart}
    \mathfrak{g} = \left\{ \left.(0_2,b)= \begin{pmatrix} 0_2 & b\\0&0\end{pmatrix}\right| b= \begin{pmatrix} b_1\\0\end{pmatrix}, b_1 \in \R  \right\}\subset \mathrm{aff}(Q).
\end{align}%
Here $0_2$ is the 2-dimensional zero matrix.
We have $\widehat{v}(H) = \widehat{(0_2,(0,s)^\top)}(H)= -s \frac{\p H}{\p s} =0$ for all $v \in \mathfrak{g}$ (see \eqref{eq:AffTraforCheck}). It follows that $I(s,\phi, p_s,p_\phi) = -p_s$ is a conserved quantity.
\end{exmp}

\begin{exmp}[Two-body problem]\label{ex:TwoBody}
Another example is a point mass orbiting a fixed point, e.g., a satellite rotating about a planet. %
We consider Cartesian coordinates with the origin of the coordinate frame in the center of the bigger point mass (e.g., the planet). %
The true Hamiltonian is given by
\begin{align}\label{eq:Kepler_Hamiltonian}
H(x,y,p_x,p_y) = \frac{1}{2}p^\top M^{-1}p - \frac{k}{\|q\|},
\end{align}
with the position of the small mass (satellite) $q^\top=(x, y)$, the corresponding conjugate momenta $p^\top=(p_x, p_y)$, the mass matrix $M=m\cdot I_2$, and the gravitation constant $k$. %
In our numerical examples $m=1$ and $k\approx1016.895$. %

Rotations around the origin, i.e.\ $(q,p) \mapsto (A^\top q,A^\top p)$ for a matrix $A \in SO(2) = \{A \in \mathrm{Gl}(2,\R) \, | \, A^\top A = I_2, \det(A)=1 \}$, \chris{constitute symmetry transformation of the dynamical system.}
These transformations can be interpreted as actions by the following subgroup of $\mathrm{Aff}(Q)$ (see Example~\ref{ex:affineLinear}):
\begin{align*}
G= & \left\{(A,0)=\left. \begin{pmatrix}A & 0\\ 0&1 \end{pmatrix} \right| A \in \R^{2 \times 2}, A^\top A = I_2, \det(A)=1 \right\}\\
& \subset \mathrm{Aff}(Q).
\end{align*}
Its Lie algebra $\mathfrak{g}$ is given as the following sub-Lie algebra of $\mathrm{aff}(Q)$
\[ \mathfrak{g} = \left\{ \left.(M,0)= \begin{pmatrix} M & 0\\0&0\end{pmatrix}\right| M \in \R^{2\times 2}, M^\top = -M  \right\}\subset \mathrm{aff}(Q).\]

For \chris{a Lie algebra element} $(M,0) \in \mathfrak{g}$ the fundamental vector field is given as $\widehat{(M,0)}_{q,p}=(-Mq,M^\top p)$ and we have $\widehat{(M,0)}_{q,p}(H)=(-Mq)^\top\nabla_qH(q,p)+(M^\top p)^\top\nabla_pH(q,p)=0$. This implies that $I(q,p)=-p^\top M q$ is a conserved quantity for the dynamics for all $M \in \mathrm{so}(2)$. Here, $\mathrm{so}(2) \subset \R^{2 \times 2}$ is the Lie algebra to $SO(2)$. It consists of all skew symmetric $2 \times 2$ matrices.
\end{exmp}

\subsection{Deep Learning for structure-preserving system identification}

Deep learning methods are widely used for data-driven identification tasks. %
We introduce the basic concepts needed for our framework and refer to the textbook \cite{Goodfellow-et-al-2016} for further information. %

A\chris{n artificial neural network is a parametrised function. %
Its structure is motivated by biological neural networks. %
Mathematically, a neural network is a composition of affine linear transformations, nonlinear functions (activation functions), and summation operators. %
The exact structure can be visualised as a directed network: starting from a specific set of nodes (input layer) values are passed through the network until they reach another set of nodes (output layer). %
The nodes are called neurons and represent summation operations, while the directed edges correspond to an application of a non-linear map (activation function) composed with an affine linear transformation. %
While the activation function is fixed, the coefficients of the affine linear transformations are free parameters. %
For visualisation, the neurons are organised in layers such that information flows from layer to layer. }

For feedforward networks, the connections only \chris{point} in the direction of the output layer, i.e.\ no backward connection and no connection within a layer are allowed. %
\chris{Universal approximation theorems for this function class can be found in} \cite[]{hornik_multilayer_1989}\chris{, for instance}. %
The network parameters $\theta$ are fitted during the learning process to minimize a user defined \emph{loss function}, usually using a variant of \chris{the} gradient descent method. %
\chris{The loss function assesses the performance of a model, for instance by calculating the average difference between model predictions and observations (mean-squared error -- MSE).}

\chris{In addition to} data fitting, 
%
physics-\chris{informed neural networks} (as proposed by \cite{raissi_physics-informed_2019}) 
enhance neural networks with physical properties: %
\chris{the solution of a PDE (encoding physical laws) is modelled as a neural network. The parameters of the neural network are sought such that the neural network fulfills the PDE and interpolates observational data simultaneously. Additionally, unknown parameters of the PDE can be fitted during the optimisation process.}

\chris{While the aforementioned work focuses on utilising prior physical knowledge to learn solutions to dynamical systems, other approaches consider learning dynamical systems that exhibit geometric properties.}
\eva{Lagrangian neural networks (LNN) by \cite{cranmer_lagrangian_2020} and Hamiltonian neural networks (HNN) by \cite{greydanus_hamiltonian_2019,bertalan19On} \chris{constitute approaches to learn models of} Lagrangian or Hamiltonian system, \chris{respectively}.}
The Lagrangian or Hamiltonian function of the system \chris{is learned} from data without assumptions on the specific form of the function. This needs to be contrasted to learning the vector field or flow map of the system directly, which does not enforce physical structure on the learned system.

Our proposed approach is based on HNN, which learns a Hamiltonian $H \colon \mathcal{M} \to \R$ modeled as a neural network based on observations $(\dot{z}^{(k)})_{k=1}^N$ of a Hamiltonian vector field $X_H \in \mathfrak{X}(\mathcal{M})$ at positions $(z^{(k)})_{k=1}^N \subset \mathcal{M}$. %
The loss function involves the error of \chris{Hamilton's} equation (see \eqref{eq:HamEQ}) for all observations, %
\begin{align}\label{eq:loss_dyn}
\ell_{\mathrm{dynamics}}
= \sum_{k=1}^{N} \| \dot{z}^{(k)} - X_H(z^{(k)}) \|_{T\mathcal{M}}^2.    
\end{align}

The derivatives to obtain $X_H(z^{(k)})$ are computed using algorithmic differentiation \cite[see][]{griewank2008evaluating}.
Further details \chris{on the optimisation algorithm and the neural network structure} are \chris{specified for each} numerical example in Section~\ref{sec:NetArchitectureHyperParams} \chris{individually}.

Based on the HNN approach, \cite{dierkes2021learning} propose to \chris{exploit prior knowledge of} symmetr\chris{ies} of the \chris{dynamical} system \chris{when} learning the Hamiltonian \chris{from data}. %
Other works, such as \cite{lishkova_discrete_2022}, are learning an underlying symmetry group alongside a discrete Lagrangian \chris{from snapshot data of trajectories}. %

\section{Simultaneous learning of symmetry actions and Hamiltonians}

Assume we are faced with an identification problem of a dynamical system known to be Hamiltonian and to possess symmetr\chris{ies}. \chris{Neither the Hamiltonian nor the symmetries or conservation laws are known explicitly but need to be learned from data.}

For a symplectic action of a group $G$ on the phase space (such as actions by affine linear transformations, for instance) we propose to learn a basis $v^{(1)},\ldots,v^{(K)}$ of a subspace $V$ of the Lie algebra $\mathfrak{g}$ such that actions by elements $g \in \exp(V)$ are symmetries of the dynamical system. The basis elements $v^{(1)},\ldots,v^{(K)}$ are learned simultaneously with the Hamiltonian $H$ from observations of the vector field. %
Here $K$ is an estimation for the dimension of the subgroup of $G$ which acts by symmetries. %
One may start with $K=1$ (or $K=0$) and successively increase $K$ until either $K=\dim(G)$ or the minimisation of the loss function introduced in the following fails to fit another symmetry. %
An identification of $V$ also reveals conserved quantities of the dynamical system as outlined in Section~\ref{sec:SymmetriesConservedQuantities}.


\subsection{Loss function for symmetry terms}


Let $\mathcal{M}^{\mathrm{o}} \subset \mathcal{M}$ be a section of the phase space $\mathcal{M}$ containing the parts of interest. $\mathcal{M}^{\mathrm{o}}$ is required to be topologically open and its closure to be a compact subset of $\mathcal{M}$.
Let $\d\mathrm{vol}$ denote the volume form on $\mathcal{M}$, i.e.\ $\d\mathrm{vol} = \d q^1 \wedge \d p_1 \wedge \ldots \wedge \d q^n \wedge \d p_n$ if $\mathcal{M}$ is represented as an open subset of $\R^{2n} \cong T^\ast \R^n$.

For $k=1,\ldots, K$ the symmetry loss term for each basis element $v^{(k)}$ is defined by
\begin{equation}\label{eq:ellsym}
\ell_{\mathrm{sym}}^{(k)} = \frac{1}{ \d \mathrm{vol}(\mathcal{M}^{\mathrm{o}})\|v^{(k)}\| } \int_{\mathcal{M}^{\mathrm{o}}} | \widehat{v}^{(k)}(H)|^2 \d \mathrm{vol}.
\end{equation}
Here, $\widehat{v}^{(k)}$ denotes the invariant vector field to $v^{(k)}$.
The term $\ell_{\mathrm{sym}}^{(k)}$ measures how invariant $H$ is under actions with group elements of $\exp({t v^{(k)} } | t \in \R)$.
The factor $\frac{1}{ \d \mathrm{vol}(\mathcal{M}^{\mathrm{o}})\|v\| }$ is a normalisation \chris{factor inversely proportional to} the volume of $\mathcal{M}^{\mathrm{o}}$ and the norm of $v$.
Equip $\mathfrak{g}$ with an inner product $\langle \cdot , \cdot \rangle $ and norm $\| \cdot \| $.
Given weights $\alpha^{(k)},\beta^{(k)} >0$ the combined symmetry loss function is
\begin{align}\label{eq:l_sym}
\ell_{\mathrm{sym}} = \sum_{k=1}^K \left(\ell_{\mathrm{sym}}^{(k)}+ \alpha^{(k)} | \| v^{(k)}\| -1 |^2
+ \beta^{(k)} \sum_{s=1}^{k-1} \langle v^{(k)},v^{(s)} \rangle\eva{^2} \right).
\end{align}

The last two terms of $\ell_{\mathrm{sym}}$ measure the orthonormality of the spanning set $v^{(1)},\ldots,v^{(K)}$ while $\sum_{k=1}^K \ell_{\mathrm{sym}}^{(k)}$ measures how well infinitesimal actions by elements of $V=\mathrm{span}(v^{(1)},\ldots,v^{(K)})$ preserve $H$.


\begin{exmp}
If we look for a 1-dimensional subgroup of the group of affine linear transformations as in Example~\ref{ex:affineLinear}, then for the Lie algebra element $v^{(1)}=(M^{(1)},b^{(1)})\in\mathfrak{g}$ we have
\begin{equation*}
\begin{split}
\ell_{\mathrm{sym}}
&=\ell_{\mathrm{sym}}^{(1)} + \alpha^{(1)} | \|v^{(1)} \|_{\mathrm{aff}(Q)} -1|^2\\
&=\ell_{\mathrm{sym}}^{(1)} + \alpha^{(1)} | \|M^{(1)}\|_\mathrm{F}+\|b^{(1)}\|_2 -1|^2    
\end{split}
\end{equation*}
with Frobenius norm $\|M^{(1)}\|_F$ (see Example~\ref{ex:affineLinear}) and Eucledian norm $\|b^{(1)}\|_2$ and with
\begin{equation}\begin{split}\label{eq:afflinearLSymm}
\ell_{\mathrm{sym}}^{(1)}
=\Theta
\int_{\mathcal{M}^{\mathrm{o}}}  &\left|(-M^{(1)}q-b^{(1)})^\top\nabla_q H(q,p)\right.\\
& \left. ~+ ({M^{(1)}}^\top p)^\top\nabla_p H(q,p)\right| ^2 \d q \d p,\end{split}
\end{equation}
with normalisation factor $\Theta = (\d \mathrm{vol}(\mathcal{M}^{\mathrm{o}}) \|v^{(1)}\|_{\mathrm{aff}(Q)})^{-1}$ and with $\d q \d p =\d q^1 \ldots \d q^n \d p_1 \ldots \d p_n$. The elements of $M^{(1)} \in \R^{n\times n}$ and $b^{(1)} \in \R^n$ are learnable parameters.

If we look for a 2-dimensional subgroup, then
\begin{equation}\label{eq:afflinearLSymm2d}
\begin{split}
\ell_{\mathrm{sym}}
&=\ell_{\mathrm{sym}}^{(1)}
+\ell_{\mathrm{sym}}^{(2)}\\
&+\alpha^{(1)}  | \|M^{(1)}\|_\mathrm{F}+\|b^{(1)}\|_2 -1|^2\\
&+ \alpha^{(2)}  | \|M^{(2)}\|_\mathrm{F}+\|b^{(2)}\|_2 -1|^2\\
&+ \beta^{(2)} (\langle M^{(2)},M^{(1)} \rangle 
+ \langle b^{(2)},b^{(1)} \rangle).
\end{split}
\end{equation}

Here $\langle M^{(2)},M^{(1)} \rangle$ is the Frobenius inner product of the matrices $M^{(2)},M^{(1)}$  (see Example~\ref{ex:affineLinear}) and $\langle b^{(2)},b^{(1)} \rangle$ the Eucledian inner product. $\alpha^{(1)},\alpha^{(2)},\beta^{(2)}>0$ are weights.
The symmetry loss $\ell_{\mathrm{sym}}^{(2)}$ is defined in analogy to $\ell_{\mathrm{sym}}^{(1)}$. The learnable symmetry parameters are given as $M^{(1)},M^{(2)}\in \R^{n \times n}$ and $b^{(1)},b^{(2)}\in \R^{n}$.

\end{exmp} 

\subsection{Training}\label{sec:training}

The Hamiltonian $H$ and the spanning set $v^{(1)},\ldots,v^{(K)}$ can now be learned using the combined loss function 
\begin{align}\label{eq:totalLoss}
    \ell = \ell_{\mathrm{dynamics}} + \chris{\delta}~\ell_{\mathrm{sym}},
\end{align} 
where $H$ is modeled as a neural network, $v^{(1)},\ldots,v^{(K)}$ are additional free parameters and $\chris{\delta} \ge 0$ is a scaling weight. %
We \chris{refer to} a model \chris{with loss function} \eqref{eq:totalLoss} \chris{as} SymHNN. %

Rather than training the network and identifying all symmetry parameters at once, the training can be performed with a low value for $K$ first. %
(Our method corresponds to HNN \chris{when} $K=0$ \chris{or $\delta =0$}.) %
Then $K$ is increased, and the training is repeated using the pre-trained network and $v^{(1)},\ldots,v^{(K-1)}$ as priors. %

\eva{
Combined learning of the system's symmetry \chris{and} its Hamiltonian reduces the function space in which the Hamiltonian is \chris{sought}.}
\chris{However, in our numerical experiments we observed that it is beneficial to start the optimisation process with $\delta = 0$ and increase $\delta$ slowly after some epochs (\eva{pre-training}). Intuitively, the adaptive increase avoids that the network is rewarded for finding symmetric representations of $H$ close to the random initialisation of $v^{(k)}$ independently of the dynamical training data.}

The integral in \eqref{eq:ellsym} \chris{is} approximated by averaging the integrand over a few points in the phase space $\mathcal{M}^\mathrm{o}$, randomly drawn in each epoch of the minimization procedure with respect to a uniform probability distribution on $\mathcal{M}^\mathrm{o}$. %

\subsection{Extensions and further remarks}

\begin{rem}
As justified \chris{in} Remark~\ref{rem:BracketClosure}, when in the learning process of Section~\ref{sec:training} Lie algebra elements $v^{(1)},v^{(2)}$ have been identified, then the Lie bracket $[v^{(1)},v^{(2)}]$ can be used as a prior for $v^{(3)}$ (unless vanishing or close to zero). Moreover, in any stage of the learning process with $K \ge 2$ (higher order iterations of) Lie brackets of already identified elements $v^{(1)},\ldots,v^{(K-1)}$ can be used to obtain initial guesses for $v^{(K)}$.
\end{rem}

\begin{rem}
Under mild assumptions, for a Lie group action of a Lie group $G$ on a symplectic manifold $\mathcal{M}$ there exists a {\em momentum map} $\mu$. %
To be more precise: Denote the paring of $\mathfrak g$ and its dual space $\mathfrak{g}^\ast$ by $\langle \cdot , \cdot \rangle_{\mathrm{pair}}$. The momentum map $\mu \colon \mathcal{M} \to \mathfrak{g}^\ast$ is a smooth map defined by $\mu(0)=0$ and by the relation $X_{\langle \mu, v \rangle } = \widehat{v}$ for all $v \in \mathfrak g$, where $\widehat{v}$ denotes the fundamental vector field to $v$ and $X_{\langle \mu, v \rangle }$ the Hamiltonian vector field to the function $\langle \mu, v \rangle \colon \mathcal{M} \to \R$.

Provided that the learned Hamiltonian $H$ accurately describes the true Hamiltonian of the system and $\ell_{\mathrm{sym}}=0$ (see \eqref{eq:l_sym}) the learned Lie algebra elements $v^{(1)},\ldots,v^{(K)}$ \chris{identified} by the described procedure yield the following functionally independent conserved quantities $\mu^{(j)} = \langle \mu, v^{(j)}\rangle_{\mathrm{pair}}$ of the system's Hamiltonian motions. In the example of affine linear transformations, the expressions $\mu^{(j)}$ recover the conserved quantities presented in Example~\ref{ex:AffLinearConserved}.

\end{rem}

\begin{rem}
\chris{Certain conserved quantities, such as the Laplace-Runge-Lenz vector in Kepler's problem, cannot be obtained via Noether's theorem from a group action on the phase space constructed as a cotangent lift of any action on the configuration space. However, it can be obtained from a symplectic group action on the phase space $\mathcal{M}$ (generalised symmetry) (see \cite{Olver1986}). Indeed, our framework is not restricted to cotangent lifted actions but } generalises directly to general symplectic group actions, i.e.\ to actions which are not cotangent lifted actions.
\end{rem}

\section{Numerical results}

For the two examples, pendulum on a cart (Example~\ref{ex:PendCart}) and 2-body problem (Example~\ref{ex:TwoBody}), numerical results are presented in this section. %
We use the library \emph{PyTorch} by \cite{paszke2019pytorch} to setup and train our models. The code for the implementation and the experiments can be found on git \url{git@github.com:eva-dierkes/HNN_withSymmetries.git}. %
More details on how we \chris{compute training} data, chose hyperparemeters of the networks, and the results are discussed in the following. %
\begin{table}[b]
    \centering
    \begin{tabular}{l@{\hskip 0.4cm}p{6.5cm}}
        $H^*$ & True Hamiltonian, see \eqref{eq:Hamiltonian_PendCart}\\
        $H_\theta$ & Learned Hamiltonian\\
        $H_{\mathrm{HNN}}$ & Learned Hamiltonian by an HNN\\
        $H_{\mathrm{SymHNN}}$ & Learned Hamiltonian by a SymHNN\\
        \hline
        \multirow{2}{*}{$\gamma^*$} & True trajectory;  \eqref{eq:Hamiltonian_PendCart} integrated with the symplectic midpoint rule\\
        \multirow{2}{*}{$\gamma_\theta$} & Trajectory of a learned model integrated with the symplectic midpoint rule\\
        \multirow{2}{*}{$\gamma_{\mathrm{BaseNN}}$} & Trajectory of a learned vector field integrated with the symplectic midpoint rule\\
        \multirow{2}{*}{$\gamma_{\mathrm{HNN}}$} & Trajectory of an HNN and the symplectic midpoint rule\\
        \multirow{2}{*}{$\gamma_{\mathrm{SymHNN}}$} & Trajectory of a SymHNN and the symplectic midpoint rule\\
        \hline
        $v^*$ & True symmetry of the system, see \eqref{eq:LieAlgebra_PendCart} \\
        $v_{\mathrm{SymHNN}}$ & Learned symmetry of the system using a SymHNN\\
    \end{tabular}
    \caption{Explanation for the used symbols and abbreviations.}
    \label{tab:SymExplanation}
\end{table}
\subsection{Generating \chris{training} data}\label{ssec:TrainingData}

In order to generate data to train and test our network, we sample initial points in the area of interest, i.e.\ $\mathcal{M}^{\mathrm{o}}$, and simulate each of them with the true dynamics for a short period of time, using a fourth order Runge-Kutta scheme \citep[see][]{runge_uber_1895} with a low error tolerance of $1\mathrm{e}{-10}$. %
\chris{Numerical integration is used as an auxiliary tool to cover the area of interest $\mathcal{M}^{\mathrm{o}}$. Since only data of the Hamiltonian vector field is used for training, rather than} \eva{trajectory information}, \chris{accuracy and structure-preservation is not important in this context.} 
%
\chris{At snapshot times of the trajectories} the vector field \chris{is evaluated} to obtain \chris{data points} $(\chris{q,p},\dot{q},\dot{p})$. \chris{When stated, Gaussian noise is added to the vector field evaluation} \eva{to simulate measurement noise}. %
As our proposed model does not require trajectory input, each data set sample is an individual training/validation/testing example. %
The data set is split into \SI{70}{\percent} training, \SI{15}{\percent} validation, and \SI{15}{\percent} test data. %
The choice of $\mathcal{M}^{\mathrm{o}}$ highly depends on the application, just as the length of the trajectories, the sampling rate, and the number of trajectories. %

For comparison \chris{we train} a dense neural network (BaseNN) \chris{that models} the vector field \chris{rather than the Hamiltonian}, and an \chris{additional} HNN without symmetry information. %

A ground truth solution $\gamma^*$ is \chris{computed} by integrating \chris{the true vector field starting from a } random initial point (\chris{sampled} in the area of interest) with \chris{the} symplectic midpoint rule \citep[see][]{hairer_geometric_2006}. %
Trajectories that are obtained by symplectic integration are denoted by $\gamma$. %
Note that learned models and symmetries are denoted by a subscript $(\cdot)_\theta$ or by a label corresponding to the used modeling approach (as BaseNN, HNN, or SymHNN), and the ground truth is labeled by a superscript~$(\cdot)^*$. %
An overview of the used abbreviations and their explanation can be found in Table~\ref{tab:SymExplanation}. %

\subsection{Net architecture and choice of hyperparameters} \label{sec:NetArchitectureHyperParams}
\chris{For HNN we use a} fully connected feedforward \chris{neural} network, as proposed by \cite{greydanus_hamiltonian_2019}.
We take three layers with 256 neurons each; as an activation function, we use $\mathrm{Softplus}(x)= \log(1+\exp(x))$.

The \emph{Adam} optimizer \citep[see][]{kingma_adam_2017} is employed to train the models with an initial learning rate of $5\mathrm{e}{-3}$. %
Additionally, a \emph{reduce on plateau} learning rate scheduler is applied, which decreases the learning rate by a factor of 0.95 if the Hamiltonian loss $\ell_{\mathrm{dynamics}}$ does not improve \chris{for more than $50$ epochs} on the validation data set. %
The learning is performed for \eva{$20~000$ epochs}, but stops early if $\ell_{\mathrm{dynamics}}$ does not decrease on the validation data set \chris{for more than $10~000$ epochs}. %

For the SymHNN models, the symmetry term is smoothly added to the loss function (as \chris{explained} in Section~\ref{sec:training}). %
The first \eva{$100$ epochs} are trained without any symmetry term, then for the next \eva{$100$ epochs} the weight \chris{$\delta$} from \eqref{eq:totalLoss} is linearly increased from $0$ to $0.5$. %

\subsection{Pendulum on a cart}
\begin{figure}
    \centering
    \includegraphics{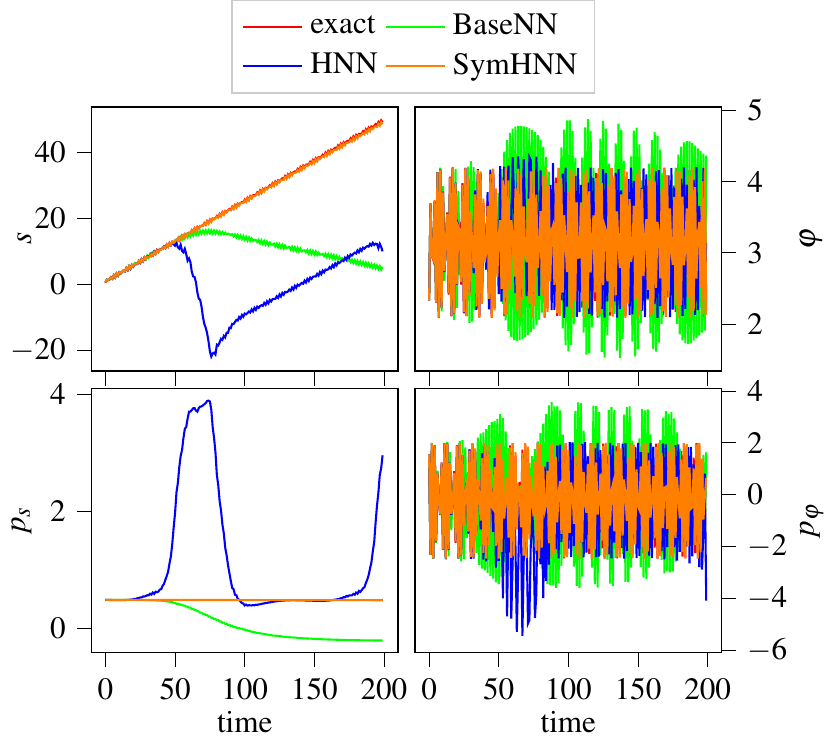}
    \caption{Computed trajectories $\gamma_\theta$ for different (learned) models for the cart-pendulum obtained with the \chris{symplectic} midpoint rule. Note: the exact trajectory is almost covered by $\gamma_{\mathrm{SymHNN}}$.}
    \label{fig:PendCart_S}
\end{figure}%
\chris{To compute} a data set for Example~\ref{ex:PendCart}, $1000$ random initial values with $|s|<5$, $|\varphi|<\pi$, $|p_s|<1$ and $|p_\varphi|<\pi$ are generated. %
Each initial value is integrated for $\SI{3}{s}$ with a sampling rate of $\SI{15}{\hertz}$. %
Only for data generation a fourth-order Runge--Kutta scheme is used to integrate the (exact) vector field induced by $H$. %
To each sample of the resulting data set, Gaussian noise with $\sigma^2=1\mathrm{e}{-2}$ is added \chris{(see \ref{ssec:TrainingData})}. %

For the training of our proposed SymHNN we set $K=1$ and  the scaling factors in \eqref{eq:l_sym}, $\alpha^{(k)}$ and $\beta^{(k)}$, are set to $1$ for all $k$. %
\begin{figure}[b]
    \centering
    \includegraphics[width=\linewidth]{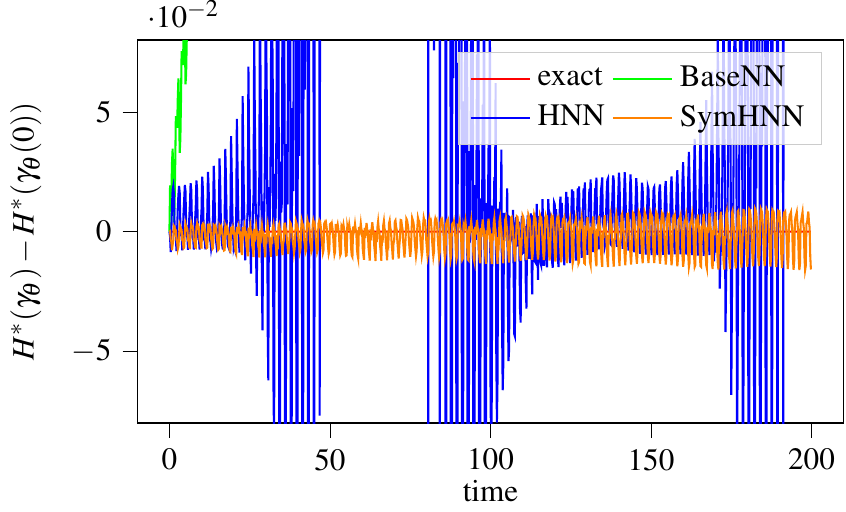}
    \caption{True cart-pendulum Hamiltonian \eqref{eq:Hamiltonian_PendCart} evaluated along the trajectories shown in Figure~\ref{fig:PendCart_S} shifted by $H^*(\gamma_\theta(0))$.}
    \label{fig:PendCart_H}
\end{figure}%
\begin{figure}
    \centering
    \includegraphics{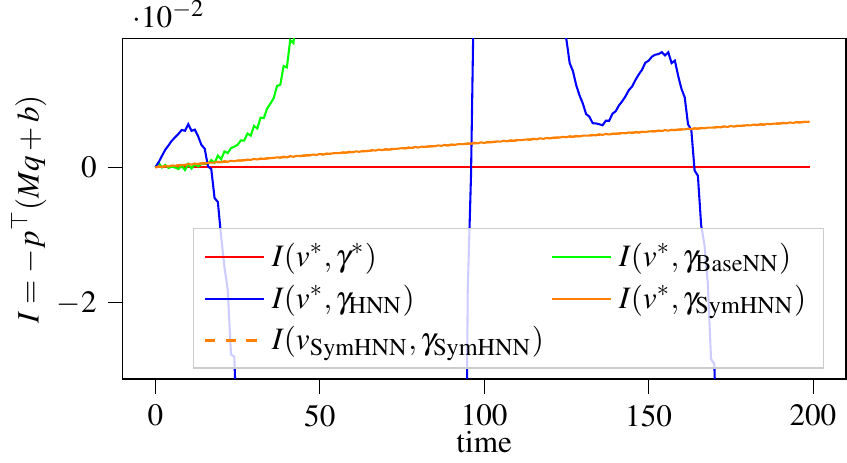}
    \caption{Conserved quantity $I$ (see \eqref{eq:ConservedAff}) evaluated \chris{along the trajectory} shown in Figure~\ref{fig:PendCart_S} (pendulum on a cart).}
    \label{fig:PendCart_ConsevedQuant}
\end{figure}%
\begin{figure}[b]
    \centering
    \includegraphics{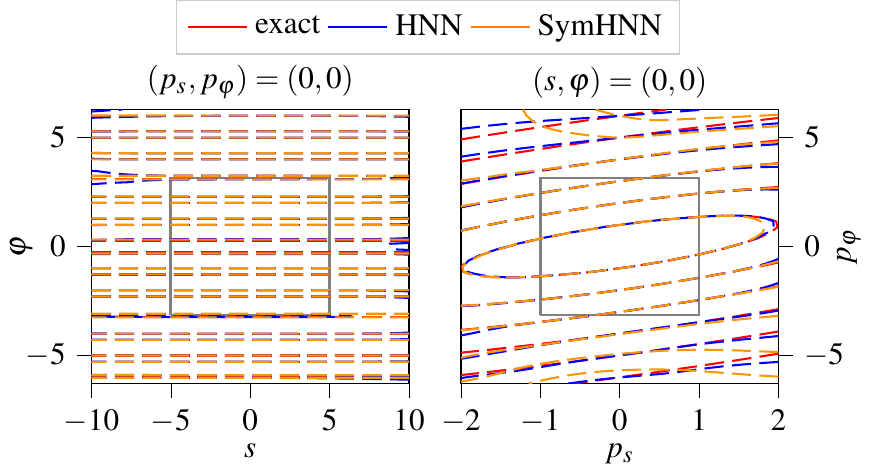}
    \caption{Level sets of (learned) cart-pendulum Hamiltonian for constant values $(p_s,p_\varphi)=(0,0)$ on the left and $(s,\varphi)=(0,0)$ on the right. The gray box indicates the \chris{training domain}.}
    \label{fig:PendCart_LevelSets}
\end{figure}%
\begin{figure*}
    \centering
    \includegraphics{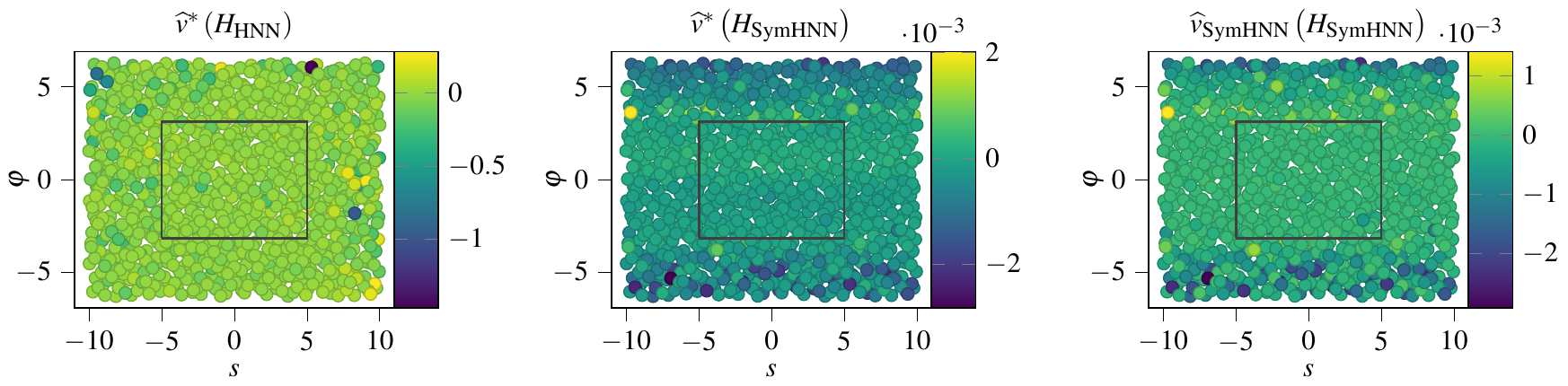}
    \caption{\chris{Symmetry error terms $\hat{v}^\ast(H_{\mathrm{HNN}})$, $\hat{v}^\ast(H_{\mathrm{SymHNN}})$, and $\hat{v}_{\mathrm{SymHNN}}^\ast(H_{\mathrm{SymHNN}})$} evaluated \chris{at} \si{1000} randomly sampled points \chris{of the phase space of the} cart pendulum projected onto the $q$-plane. \chris{The $p$-values are drawn with $|p_s|<2$ and $|p_\varphi|<2 \pi$.} Notice that the scaling of the color bars differ. \eva{The gray boxes indicate the area where the initial values for the training \chris{data} are sampled.}}
    \label{fig:PendCart_vHat_overQ}
\end{figure*}%

\eva{
\setlength{\tabcolsep}{5pt}
\begin{table}[t]
    \centering
    \begin{tabular}{ll|c|c|c}
         &Loss  & train & validation & test \\
         \hline
         BaseNN &$\ell_{\mathrm{VF}}$& $0.00296$ & $ 0.00351$ & $0.00387$\\
         HNN & $\ell_{\mathrm{dynamics}}$& $0.00305$ & $0.00332$& $0.00363$\\
         SymHNN & $\ell$& $0.00308$ & $0.00324$& $0.00342$\\
         &$\ell_{\mathrm{dynamics}}$& $0.00307$ & $0.00323$ & $0.00340$\\
         &$\ell_{\mathrm{sym}}$& \multicolumn{3}{c}{\rule[1mm]{1cm}{.4pt}  $1.642\mathrm{e}{-6}$  \rule[1mm]{1cm}{.4pt}} \\
    \end{tabular}
    \caption{Train, validation and test losses for the learned models for the pendulum on a cart example. $\ell_{\mathrm{VF}}$ is the MSE between learned and true vector field for BaseNN. Note that $\ell_{\mathrm{sym}}$ is not evaluated on training data but on a mesh in $\mathcal{M}^\circ$.}
    \label{tab:PendCart_losses}
\end{table}
}
\eva{Table~\ref{tab:PendCart_losses} shows the training, validation and test losses for the trained models. %
For this, we evaluate the loss functions $\ell_{\mathrm{dynamic}}$ \eqref{eq:loss_dyn}, $\ell_{\mathrm{sym}}$\eqref{eq:l_sym} and $\ell$ \eqref{eq:totalLoss} \chris{on the data sets produced in \ref{ssec:TrainingData}}. %
For the BaseNN model we compute the MSE between the learned and the true vector field on the data set. %
For the SymHNN model $\ell_{\mathrm{sym}}$ is evaluated over a fixed grid on $\mathcal{M}^\circ$ independent of the training data set. %
The values of $\ell_{\mathrm{dynamics}}$ are approximately the same for HNN and SymHNN. 
This indicates that the introduction of an additional symmetry term neither harms nor benefits the minimisation of $\ell_{\mathrm{dynamics}}$ during training. }%
Figure~\ref{fig:PendCart_S} \chris{shows the long-time behaviour of trajectories with } initial \chris{values} \mbox{$[q,p]=[0,  2.34, 0.49, 1.54]$} \chris{for BaseNN, HNN, and SymHNN}. %
The true Hamiltonian~\eqref{eq:Hamiltonian_PendCart} is denoted by $H^*$.
\chris{Figure~\ref{fig:PendCart_H} shows the conservation of $H^*$ along the trajectories of Figure~\ref{fig:PendCart_S}. }%
Notice that since system dynamics depend on derivatives of $H$ only, constant offsets \chris{in the values of $H$ are} irrelevant. %
The evaluation of $H$ for the SymHNN trajectory stays almost constant and close to the reference.
The Hamiltonian for the BaseNN grows quickly and for the HNN, it oscillates with a high amplitude. %

It should be highlighted that SymHNN \chris{identifies the symmetry parameters highly accurately}: \chris{it learns} $b^{(1)}=[1.0000,-1.0348\mathrm{e}{-05}]$ and 
\[
M^{(1)}=\begin{pmatrix}-1.21\mathrm{e}{-6} & -1.91\mathrm{e}{-6}\\\-9.30\mathrm{e}{-9} & -4.22\mathrm{e}{-6}\end{pmatrix},
\]
which is close to $[1,0]$ and the zero matrix $0_2$, respectively. %
Since this example is known to be invariant in the position~$s$, the momentum $p_s$ is conserved, which leads to a constant $p_s$ for each trajectory. %
In Figure~\ref{fig:PendCart_S} it can be seen that the momentum $p_s$ for the BaseNN and the HNN are not conserved, whereas the momentum for our proposed SymHNN is preserved \chris{to high accuracy}. %
To confirm this, the values for $\ell_{\mathrm{sym}}^{(k)}$ are computed with $v^*$ and $1000$ randomly sampled points (in a \chris{domain}, which is \chris{larger} than the training \chris{domain}). %
An improvement by a factor of $100$ is observed (HNN: $3.1478$ and SymHNN: $0.0136$). %
The value for the SymHNN model with its learned value $v_{SymHNN}$ is $0.0121$, which indicates that the model preserves the learned symmetry slightly better than the true \chris{symmetry}. %
However, since the integration in \eqref{eq:ellsym} is not performed exactly and the symmetry losses $\ell_{\mathrm{sym}}^{(k)}$ are not precisely zero, the conjugate momentum $p_s$ is not exactly but only approximately conserved. %

Figure~\ref{fig:PendCart_ConsevedQuant} shows the conserved quantities \chris{corresponding to }the true symmetry $v^*$ or the learned symmetry $v_{\mathrm{SymHNN}}$ (see Example~\ref{ex:AffLinearConserved}) evaluated \chris{along} the trajectories shown in Figure~\ref{fig:PendCart_S}.
$I$ increases quickly for the BaseNN model and oscillates and slightly drifts away for the HNN, whereas for the SymHNN it only slightly drifts away but has no high oscillations. %
\begin{figure}[b]
    \centering
    \includegraphics{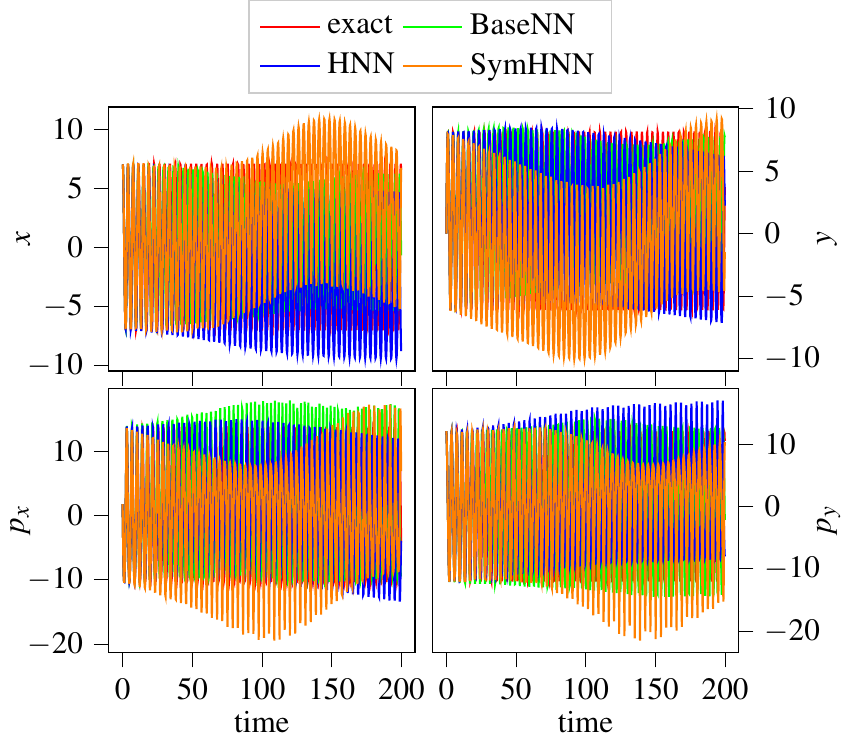}
    \caption{Trajectories of different (learned) models for the cartesian two-body problem.}
    \label{fig:KeplerCartesian_S}
\end{figure}
The level sets of the Hamiltonian are shown in Figure~\ref{fig:PendCart_LevelSets} for constant values $(p_s,p_\varphi)=(0,0)$ on the left and for $(s,\varphi)=(0,0)$ on the right. %
Since for BaseNN no Hamiltonian is learned, it is not shown in this figure. %
The gray box indicates the \chris{domain of training data}. %
The level sets inside the box match the true ones rather well for both the HNN and the SymHNN model. %
Outside the gray box, both models start deviating from the true solution. %
Together with the trajectories from Figure~\ref{fig:PendCart_S}, we conclude that conservation of $p_s$ generalizes outside the \chris{training domain}.
\chris{Figure~\ref{fig:PendCart_vHat_overQ} shows the Lie derivatives $\hat{v}^\ast(H_{\mathrm{HNN}})$, $\hat{v}^\ast(H_{\mathrm{SymHNN}})$, and $\hat{v}_{\mathrm{SymHNN}}^\ast(H_{\mathrm{SymHNN}})$ of the learned Hamiltonians $H_{\mathrm{HNN}}$ and $H_{\mathrm{SymHNN}}$ along the fundamental vector fields $\hat{v}^\ast$ or $\hat{v}_{\mathrm{SymHNN}}$ that correspond to the true symmetry of the cart-pendulum model or the learned symmetry, respectively. This provides a measures how symmetric the learned model is with respect to the true symmetry and how consistent SymHNN is with the identified symmetry. The Lie derivatives $\hat{v}^\ast(H_{\mathrm{HNN}})$, $\hat{v}^\ast(H_{\mathrm{SymHNN}})$ constitute functions on the phase space. Values close to zero indicate symmetry.}
The values of \chris{these functions are} plotted \chris{over} $1000$ \chris{randomly sampled phase space elements}. %
It can be observed that, overall, the symmetry error is smaller for the SymHNN models than for HNN. \chris{As expected, $\hat{v}_\mathrm{SymHNN}$ describes the symmetry in the SymHNN model most accurately. } 
\begin{figure}
    \centering
    \includegraphics{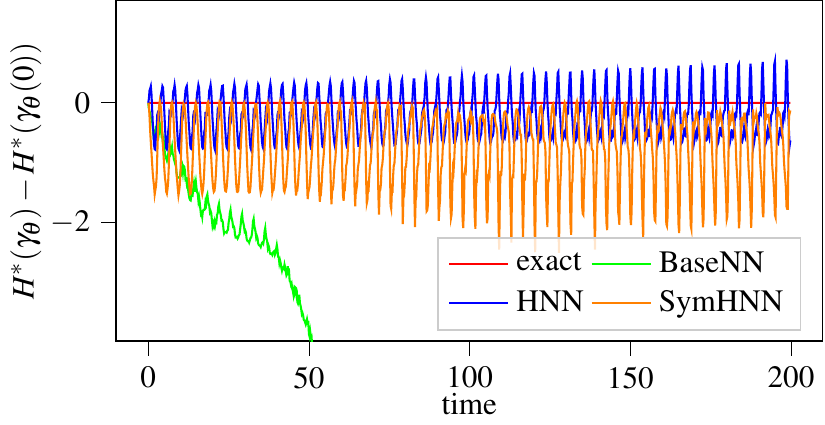}
    \caption{True two-body Hamiltonian \eqref{eq:Kepler_Hamiltonian} evaluated along the trajectories shown in Figure~\ref{fig:KeplerCartesian_S}  shifted by $H^*(\gamma_\theta(0))$.}
    \label{fig:KeplerCartesian_H}
\end{figure}
\subsection{Two-body problem in cartesian coordinates}

\begin{figure*}[t]
    \centering
    \includegraphics{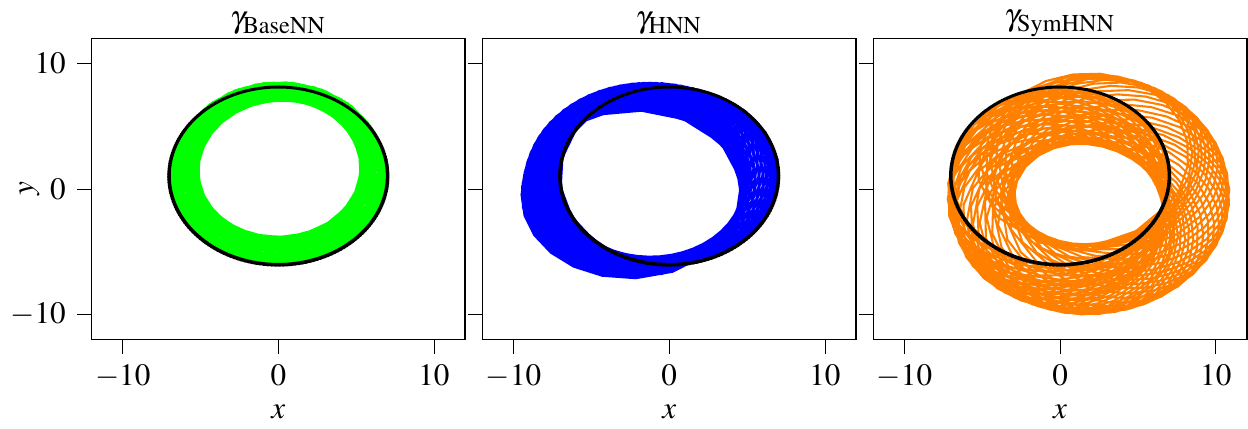}
    \caption{Two-body states shown in Figure~\ref{fig:KeplerCartesian_S} projected on $(x,y)$-plane. \eva{The black circles show the projection of the true trajectory.}}
    \label{fig:KeplerCartesian_S_phaseplot}
\end{figure*}
\begin{figure*}
    \centering
    \includegraphics{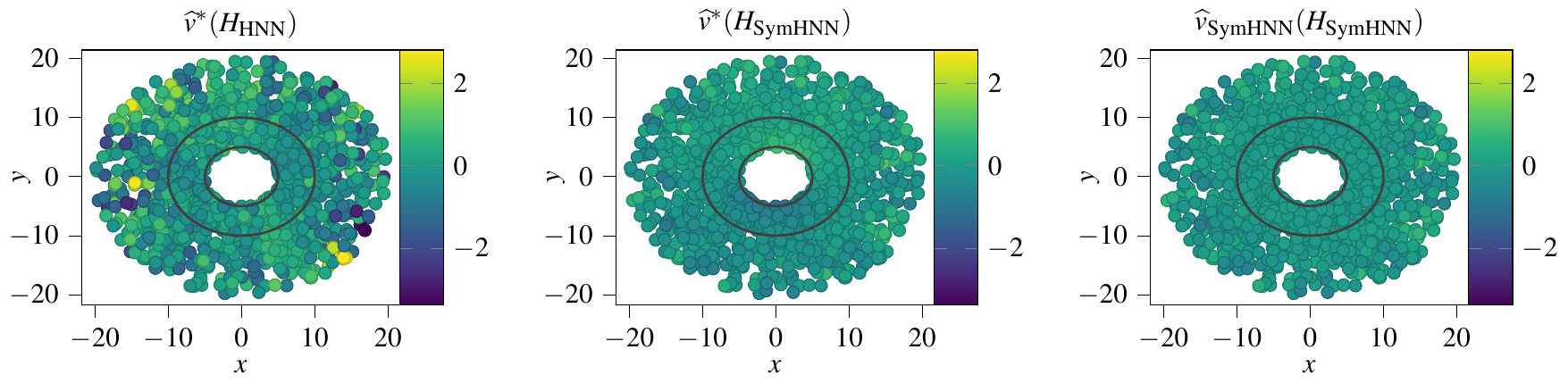}
    \caption{\chris{The directional derivatives $\widehat{v}(H_\theta)$ in the two-body problem are evaluated on \si{1000} points of the phase space (randomly sampled on the training domain with radially extended range of $q=(x,y)$ values) and plotted over the $q$-plane (ignoring the $p$-components). The values for HNN are plotted to the left and for SymHNN in the centre figure. While these plots correspond to the true symmetry $\widehat{v}^\ast$, the plot to the right shows the corresponding plot for SymHNN with the learned symmetry $\widehat{v}_{\mathrm{SymHNN}}$.}  \eva{The area enclosed by the two gray circles shows the training data domain projected to the $q$-plane.}
    }
    \label{fig:KeplerCartesian_vHat_confetti}
\end{figure*}
\begin{figure}[b]
    \centering
    \includegraphics{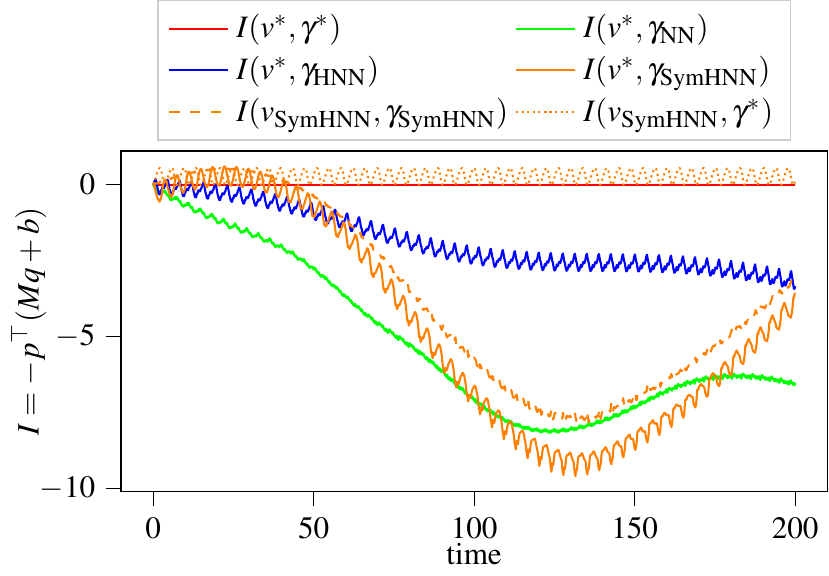}
    \caption{Conserved quantity $I$ (see \eqref{eq:ConservedAff}) evaluated for states of the two-body problem shown in Figure~\ref{fig:KeplerCartesian_S}.}
    \label{fig:KeplerCartesian_ConservedQuant}
\end{figure}

\begin{figure*}
    \centering
    \includegraphics{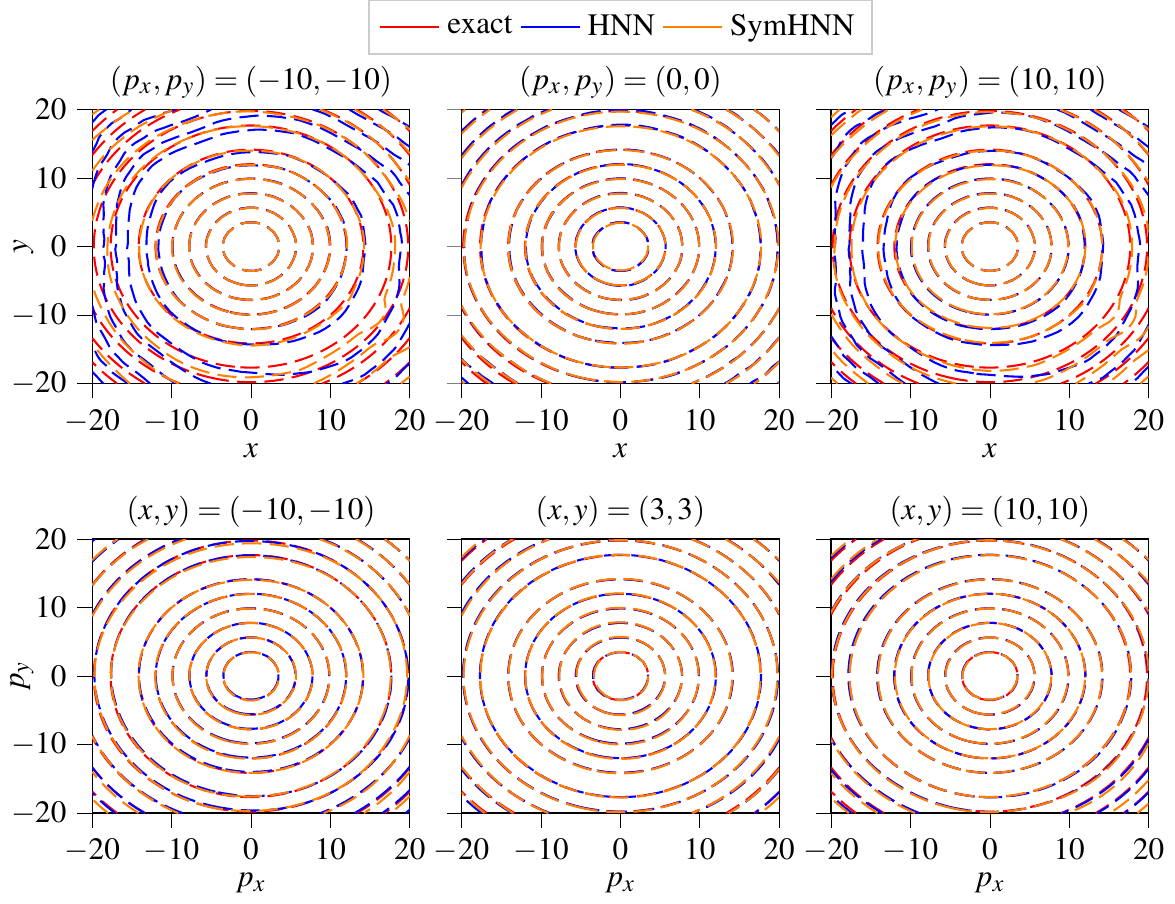}
    \caption{Level sets of the different learned two-body Hamiltonian for fixed values for some coordinates (as indicated by the respective titles).}
    \label{fig:KeplerCartesian_HLevelSets}
\end{figure*}
\begin{figure}[b]
    \centering
    \includegraphics{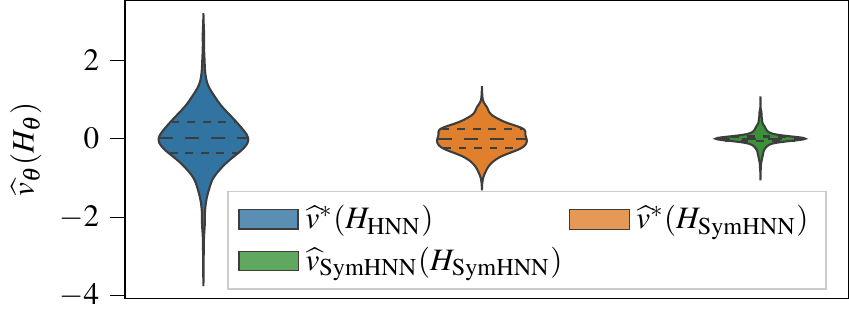}
    \caption{Distribution of the values of $\widehat{v}\chris{(H_\theta)}$ shown in Figure~\ref{fig:KeplerCartesian_vHat_confetti}.}
    \label{fig:KeplerCartesian_vHat_violine}
\end{figure}
In order to generate training data for the \chris{planar} two-body problem (Example \ref{ex:TwoBody}), points \chris{within} a radius between $5$ and $10$ \chris{of the origin} are sampled. The momentum $p$ is constructed to be orthogonal to $q$ with norm $\|p\|=(1\pm 0.3)\left(\sqrt{\frac{k}{\|q\|}}\right)$. \chris{Both orientations are considered. }%
\chris{Notice that} {$\|p\|=\sqrt{\frac{k}{\|q\|}}$} \chris{would} lead to circular trajectories \chris{because} the \emph{vis viva} equation \chris{is fulfilled for these values} \cite[see][]{vallado2001fundamentals}. %

$5000$ trajectories of $\SI{10}{s}$ length are generated using a fourth-order Runge--Kutta method with a sampling rate of $\SI{1}{\hertz}$. \chris{At each snapshot the Hamiltonian vector field to \eqref{eq:Kepler_Hamiltonian} is evaluated} and Gaussian noise with $\sigma^2=1\mathrm{e}{-2}$ is added. %
The constructed data set is split into training (\SI{70}{\percent}), validation (\SI{15}{\percent}) and test data (\SI{15}{\percent}). %

\eva{
\setlength{\tabcolsep}{5pt}
\begin{table}[b]
    \centering
    \begin{tabular}{ll|c|c|c}
         &Loss  & train & validation & test \\
         \hline
         BaseNN &$\ell_{\mathrm{VF}}$& $0.00668$ & $0.04591$ & $0.03034$\\
         HNN &$\ell_{\mathrm{dynamics}}$& $0.01411$ & $0.01651$& $0.01895$\\
         SymHNN &$\ell$& $0.08746 $ & $0.13644$& $0.10756$\\
         &$\ell_{\mathrm{dynamics}}$& $0.01470$ & $0.06368$ & $0.03480$\\
         &$\ell_{\mathrm{sym}}$& \multicolumn{3}{c}{\rule[1mm]{1cm}{.4pt}  $0.14553$  \rule[1mm]{1cm}{.4pt}} \\
    \end{tabular}
    \caption{Train, validation and test losses for the different \chris{trained} models for the two-body problem. $\ell_{\mathrm{VF}}$ is the MSE \chris{between} learned and true vector field for BaseNN. Note that $\ell_{\mathrm{sym}}$ is not evaluated on training data but on \chris{a mesh in} $\mathcal{M}^\circ$.}
    \label{tab:Kepler_losses}
\end{table}
}
\eva{
Table~\ref{tab:Kepler_losses} shows the training, validation and test loss function values for the trained models, similar to Table~\ref{tab:PendCart_losses}. %
The scaling parameter is $\delta=0.5$ (see \eqref{eq:totalLoss}). 
}

Figure~\ref{fig:KeplerCartesian_S} shows the longtime trajectories $\gamma_\theta$ that are obtained using the learned models and integration using the implicit midpoint rule for the initial point $[7.0,0.0,1.72,12.05]$. %
None of the models \chris{recovers the true solution accurately}. %

Figure~\ref{fig:KeplerCartesian_H} \chris{shows the conservation of the} true Hamiltonian $H^*$, see \eqref{eq:Kepler_Hamiltonian}, along the trajectories from Figure~\ref{fig:KeplerCartesian_S}. 
The Hamiltonian \chris{evaluated along the numerically computed trajectory of} the exact model oscillates with an amplitude of about $\mathrm{e}{-5}$ (\chris{The oscillation is} not visible in Figure~\ref{fig:KeplerCartesian_H}). %
\chris{This indicates that any energy error due to numerical integration with the symplectic midpoint rule is negligible.} %
The energy for the BaseNN \chris{decreases very quickly}, slightly swings up for the HNN and, besides the oscillations, stays at the same level for the SymHNN. 

The conserved quantities, recall \eqref{eq:ConservedAff}, are shown in Figure~\ref{fig:KeplerCartesian_ConservedQuant} for the different combinations of trajectories and symmetries. %
The solid lines show the conserved quantity for the true symmetry, the dashed orange line shows the conserved quantity for the SymHNN trajectory with the learned symmetry and the dotted line shows the true trajectory with the learned symmetry $v_{\mathrm{SymHNN}}$. %
The BaseNN and the HNN are not capable of conserving $I$, whereas for SymHNN the conserved quantity oscillates. %
The orange dotted line indicates how \chris{well the true trajectory preserves the learned conserved quantity.} %
It only oscillates with a small amplitude. %

The projection from the trajectories from Figure~\ref{fig:KeplerCartesian_S} onto the $(x,y)$-plane are shown in Figure~\ref{fig:KeplerCartesian_S_phaseplot}\eva{, as well as the exact orbit for reference. }%
The exact system is completely integrable, i.e.\ the degrees of freedom equate to the number of independent, commuting conserved quantities. Compact level sets of the conserved quantities form topological tori (Liouville tori) and Hamiltonian motions are quasi-periodic motions on these tori \citep{Libermann1987}.
The regularity in the plots corresponding to HNN and SymHNN suggest that the motions of these models are (almost) quasi-periodic motions on some modified tori in the phase space. This suggests that a completely integrable structure is approximately present in the learned models. In contrast, the unstructured motions exhibited by BaseNN appear not to \chris{be governed by} completely integrable dynamics.


In Figure~\ref{fig:KeplerCartesian_vHat_confetti}, the values of $\widehat{v}_{(q,p)}(H_\theta)$ are shown for $1000$ randomly sampled pairs of $(q,p)$ projected on the $(x,y)$-plane\eva{, additionally, the area where the initial values for the training data are sampled is marked. }%
The samples are \chris{drawn from the a slightly larger set than the training domain}. %
\chris{The directional derivative (Lie derivative) $\widehat{v}(H_\theta)$ vanishes, if $H_\theta$ is invariant with respect to the infinitesimal action by the Lie algebra element $v$. }%
\chris{Comparing the left and the centre} figure indicates that the symmetry is better conserved by the SymHNN as not so many samples with high $\widehat{v}_{(q,p)}$-values are present. %
Furthermore, the values in the right figure are even smaller, indicating that the SymHNN model preserves the learned symmetry the best. %

Figure~\ref{fig:KeplerCartesian_vHat_violine} confirms this result: \chris{the violin plots (see \cite{ViolinPlots}, for instance) are based on} the same points as in Figure~\ref{fig:KeplerCartesian_vHat_confetti} \chris{and indicate how likely and by how much the quantities $\widehat{v}^\ast(H_\mathrm{HNN})$, $\widehat{v}^\ast(H_\mathrm{SymHNN})$, $\widehat{v}_{\mathrm {SymHNN}}(H_\mathrm{HNN})$ deviate from zero. Moreover, dashed lines indicate boundaries of quartiles. The line between the second and third quartile corresponds to the median.} %
The violin on the right \chris{has the smallest volume indicating that} the learned symmetry is approximately preserved by the SymHNN model. %

\chris{Figure~\ref{fig:KeplerCartesian_HLevelSets} shows several cross sections of} level sets of the (learned) Hamiltonian. 
The two figures in the centre show that the level sets for $(p_x,p_y)=(0,0)$ and for $(x,y)=(3,3)$ are approximated quite \chris{well} by the HNN and the SymHNN. %
Further away from the training data\eva{, as $p=\pm 10$} \chris{SymHNN performs better as can be seen in the plots in the} \eva{first and last column}.


\section{Conclusion and future work}
We propose a neural network approach for simultaneously learning a system's Hamiltonian and its symmetries based on snapshot data \chris{of the vector field}. %
To preserve the symplectic structure encoded in the data, the NN is trained to learn a Hamiltonian, since the vector field can then be generated via automatic differentiation \citep[cf.][]{greydanus_hamiltonian_2019,bertalan19On,dierkes2021learning}.
Extending this approach, known as the HNN method, we simultaneously identify inherent symmetries. %
For this, we make an ansatz for a symplectic Lie-group action on the phase space (e.g., affine linear transformations) and identify a subgroup whose \chris{infinitesimal} actions leave the Hamiltonian invariant. %
The identified symmetries then correspond to conserved quantities of the Hamiltonian motion, such as conserved momentum in the studied cart-pendulum example. %

The presented numerical examples demonstrate that the proposed framework is capable of learning a Hamiltonian and to identify symmetries of the system. %
Moreover, incorporation of the learned symmetries into the learned model improves qualitative aspects of the predicted motions.

A discrete-time variant of this approach might directly learn modified Hamiltonian and discrete-time symmetries \citep[cf.][]{SSI,LSI}.
Symmetry in Hamiltonian systems give rise to relative equilibria, to which the learning framework can be extended in future. %
\eva{Additionally, the proposed SympNets by \cite{SympNets} are learning symplectic discrete flows of the system where a combination might be beneficial. }%

\eva{The proposed framework may be of interest to various applications where exact models are unknown or too complicated to be used in numerical simulations. %
Among others, to approximate trim primitives in autonomous driving applications as in \cite{pedrosa22learning}, to model smart materials as in \cite{baltes2022a}, or to learn dynamical systems from high dimensional data (e.g.\ from video data) as proposed in \cite{bertalan19On,mason2022learning} \chris{in combination with} auto/decoder type networks. %
}

\begin{acknowledgments}
E.~Dierkes acknowledges funding by the Deutsche Forschungsgemeinschaft (DFG) {Project number 281474342.} C.~Offen acknowledges the Ministerium für Kultur und Wissenschaft des Landes Nordrhein-Westfalen.
\end{acknowledgments}

\bibliography{Bibliography.bib}                                                   

\begin{thebibliography}{38}%
\makeatletter
\providecommand \@ifxundefined [1]{%
 \@ifx{#1\undefined}
}%
\providecommand \@ifnum [1]{%
 \ifnum #1\expandafter \@firstoftwo
 \else \expandafter \@secondoftwo
 \fi
}%
\providecommand \@ifx [1]{%
 \ifx #1\expandafter \@firstoftwo
 \else \expandafter \@secondoftwo
 \fi
}%
\providecommand \natexlab [1]{#1}%
\providecommand \enquote  [1]{``#1''}%
\providecommand \bibnamefont  [1]{#1}%
\providecommand \bibfnamefont [1]{#1}%
\providecommand \citenamefont [1]{#1}%
\providecommand \href@noop [0]{\@secondoftwo}%
\providecommand \href [0]{\begingroup \@sanitize@url \@href}%
\providecommand \@href[1]{\@@startlink{#1}\@@href}%
\providecommand \@@href[1]{\endgroup#1\@@endlink}%
\providecommand \@sanitize@url [0]{\catcode `\\12\catcode `\$12\catcode
  `\&12\catcode `\#12\catcode `\^12\catcode `\_12\catcode `\%12\relax}%
\providecommand \@@startlink[1]{}%
\providecommand \@@endlink[0]{}%
\providecommand \url  [0]{\begingroup\@sanitize@url \@url }%
\providecommand \@url [1]{\endgroup\@href {#1}{\urlprefix }}%
\providecommand \urlprefix  [0]{URL }%
\providecommand \Eprint [0]{\href }%
\providecommand \doibase [0]{http://dx.doi.org/}%
\providecommand \selectlanguage [0]{\@gobble}%
\providecommand \bibinfo  [0]{\@secondoftwo}%
\providecommand \bibfield  [0]{\@secondoftwo}%
\providecommand \translation [1]{[#1]}%
\providecommand \BibitemOpen [0]{}%
\providecommand \bibitemStop [0]{}%
\providecommand \bibitemNoStop [0]{.\EOS\space}%
\providecommand \EOS [0]{\spacefactor3000\relax}%
\providecommand \BibitemShut  [1]{\csname bibitem#1\endcsname}%
\let\auto@bib@innerbib\@empty
\bibitem [{\citenamefont {Baltes}\ \emph {et~al.}(2022)\citenamefont {Baltes},
  \citenamefont {Kunze}, \citenamefont {Prechtl}, \citenamefont {Seelecke},\
  and\ \citenamefont {Rizzello}}]{baltes2022a}%
  \BibitemOpen
  \bibfield  {author} {\bibinfo {author} {\bibnamefont {Baltes}, \bibfnamefont
  {M.}}, \bibinfo {author} {\bibnamefont {Kunze}, \bibfnamefont {J.}}, \bibinfo
  {author} {\bibnamefont {Prechtl}, \bibfnamefont {J.}}, \bibinfo {author}
  {\bibnamefont {Seelecke}, \bibfnamefont {S.}}, \ and\ \bibinfo {author}
  {\bibnamefont {Rizzello}, \bibfnamefont {G.}},\ }\bibfield  {title} {\enquote
  {\bibinfo {title} {A bi-stable soft robotic bendable module driven by
  silicone dielectric elastomer actuators: design, characterization, and
  parameter study},}\ }\href {\doibase 10.1088/1361-665X/ac96df} {\bibfield
  {journal} {\bibinfo  {journal} {Smart Materials and Structures}\ }\textbf
  {\bibinfo {volume} {31}} (\bibinfo {year} {2022}),\
  10.1088/1361-665X/ac96df}\BibitemShut {NoStop}%
\bibitem [{\citenamefont {Bertalan}\ \emph {et~al.}(2019)\citenamefont
  {Bertalan}, \citenamefont {Dietrich}, \citenamefont {Mezić},\ and\
  \citenamefont {Kevrekidis}}]{bertalan19On}%
  \BibitemOpen
  \bibfield  {author} {\bibinfo {author} {\bibnamefont {Bertalan},
  \bibfnamefont {T.}}, \bibinfo {author} {\bibnamefont {Dietrich},
  \bibfnamefont {F.}}, \bibinfo {author} {\bibnamefont {Mezić}, \bibfnamefont
  {I.}}, \ and\ \bibinfo {author} {\bibnamefont {Kevrekidis}, \bibfnamefont
  {I.~G.}},\ }\bibfield  {title} {\enquote {\bibinfo {title} {On learning
  hamiltonian systems from data},}\ }\href {\doibase 10.1063/1.5128231}
  {\bibfield  {journal} {\bibinfo  {journal} {Chaos: An Interdisciplinary
  Journal of Nonlinear Science}\ }\textbf {\bibinfo {volume} {29}},\ \bibinfo
  {pages} {121107} (\bibinfo {year} {2019})},\ \Eprint
  {http://arxiv.org/abs/https://doi.org/10.1063/1.5128231}
  {https://doi.org/10.1063/1.5128231} \BibitemShut {NoStop}%
\bibitem [{\citenamefont {Bloch}(2003)}]{Bloch}%
  \BibitemOpen
  \bibfield  {author} {\bibinfo {author} {\bibnamefont {Bloch}, \bibfnamefont
  {A.~M.}},\ }\bibfield  {title} {\enquote {\bibinfo {title} {Nonholonomic
  mechanics and control},}\ }in\ \href {\doibase 10.1007/b97376_5} {\emph
  {\bibinfo {booktitle} {Nonholonomic Mechanics and Control. Interdisciplinary
  Applied Mathematics}}},\ Vol.~\bibinfo {volume} {24}\ (\bibinfo  {publisher}
  {Springer},\ \bibinfo {year} {2003})\BibitemShut {NoStop}%
\bibitem [{\citenamefont {Brunton}, \citenamefont {Proctor},\ and\
  \citenamefont {Kutz}(2016)}]{brunton_discovering_2016}%
  \BibitemOpen
  \bibfield  {author} {\bibinfo {author} {\bibnamefont {Brunton}, \bibfnamefont
  {S.~L.}}, \bibinfo {author} {\bibnamefont {Proctor}, \bibfnamefont {J.~L.}},
  \ and\ \bibinfo {author} {\bibnamefont {Kutz}, \bibfnamefont {J.~N.}},\
  }\bibfield  {title} {\enquote {\bibinfo {title} {Discovering governing
  equations from data by sparse identification of nonlinear dynamical
  systems},}\ }\href {\doibase 10.1073/pnas.1517384113} {\bibfield  {journal}
  {\bibinfo  {journal} {Proc. Natl. Acad. Sci. USA}\ }\textbf {\bibinfo
  {volume} {113}},\ \bibinfo {pages} {3932--3937} (\bibinfo {year}
  {2016})}\BibitemShut {NoStop}%
\bibitem [{\citenamefont {Chen}\ and\ \citenamefont
  {Tao}(2021)}]{chen21datadriven}%
  \BibitemOpen
  \bibfield  {author} {\bibinfo {author} {\bibnamefont {Chen}, \bibfnamefont
  {R.}}\ and\ \bibinfo {author} {\bibnamefont {Tao}, \bibfnamefont {M.}},\
  }\bibfield  {title} {\enquote {\bibinfo {title} {Data-driven prediction of
  general hamiltonian dynamics via learning exactly-symplectic maps},}\ }in\
  \href {https://proceedings.mlr.press/v139/chen21r.html} {\emph {\bibinfo
  {booktitle} {Proceedings of the 38th International Conference on Machine
  Learning}}},\ \bibinfo {series} {Proceedings of Machine Learning Research},
  Vol.\ \bibinfo {volume} {139},\ \bibinfo {editor} {edited by\ \bibinfo
  {editor} {\bibfnamefont {M.}~\bibnamefont {Meila}}\ and\ \bibinfo {editor}
  {\bibfnamefont {T.}~\bibnamefont {Zhang}}}\ (\bibinfo  {publisher} {PMLR},\
  \bibinfo {year} {2021})\ pp.\ \bibinfo {pages} {1717--1727}\BibitemShut
  {NoStop}%
\bibitem [{\citenamefont {Chen}\ \emph {et~al.}(2020)\citenamefont {Chen},
  \citenamefont {Zhang}, \citenamefont {Arjovsky},\ and\ \citenamefont
  {Bottou}}]{ChenZAB20}%
  \BibitemOpen
  \bibfield  {author} {\bibinfo {author} {\bibnamefont {Chen}, \bibfnamefont
  {Z.}}, \bibinfo {author} {\bibnamefont {Zhang}, \bibfnamefont {J.}}, \bibinfo
  {author} {\bibnamefont {Arjovsky}, \bibfnamefont {M.}}, \ and\ \bibinfo
  {author} {\bibnamefont {Bottou}, \bibfnamefont {L.}},\ }\bibfield  {title}
  {\enquote {\bibinfo {title} {Symplectic recurrent neural networks},}\ }in\
  \href {https://openreview.net/forum?id=BkgYPREtPr} {\emph {\bibinfo
  {booktitle} {8th International Conference on Learning Representations, ICLR
  2020, Addis Ababa, Ethiopia, April 26-30, 2020}}}\ (\bibinfo  {publisher}
  {OpenReview.net},\ \bibinfo {year} {2020})\BibitemShut {NoStop}%
\bibitem [{\citenamefont {Cranmer}\ \emph {et~al.}(2020)\citenamefont
  {Cranmer}, \citenamefont {Greydanus}, \citenamefont {Hoyer}, \citenamefont
  {Battaglia}, \citenamefont {Spergel},\ and\ \citenamefont
  {Ho}}]{cranmer_lagrangian_2020}%
  \BibitemOpen
  \bibfield  {author} {\bibinfo {author} {\bibnamefont {Cranmer}, \bibfnamefont
  {M.}}, \bibinfo {author} {\bibnamefont {Greydanus}, \bibfnamefont {S.}},
  \bibinfo {author} {\bibnamefont {Hoyer}, \bibfnamefont {S.}}, \bibinfo
  {author} {\bibnamefont {Battaglia}, \bibfnamefont {P.}}, \bibinfo {author}
  {\bibnamefont {Spergel}, \bibfnamefont {D.}}, \ and\ \bibinfo {author}
  {\bibnamefont {Ho}, \bibfnamefont {S.}},\ }\bibfield  {title} {\enquote
  {\bibinfo {title} {Lagrangian neural networks},}\ }in\ \href
  {https://openreview.net/forum?id=iE8tFa4Nq} {\emph {\bibinfo {booktitle}
  {ICLR 2020 Workshop on Integration of Deep Neural Models and Differential
  Equations}}}\ (\bibinfo {year} {2020})\BibitemShut {NoStop}%
\bibitem [{\citenamefont {Dehmamy}\ \emph {et~al.}(2021)\citenamefont
  {Dehmamy}, \citenamefont {Walters}, \citenamefont {Liu}, \citenamefont
  {Wang},\ and\ \citenamefont {Yu}}]{Dehmany2021}%
  \BibitemOpen
  \bibfield  {author} {\bibinfo {author} {\bibnamefont {Dehmamy}, \bibfnamefont
  {N.}}, \bibinfo {author} {\bibnamefont {Walters}, \bibfnamefont {R.}},
  \bibinfo {author} {\bibnamefont {Liu}, \bibfnamefont {Y.}}, \bibinfo {author}
  {\bibnamefont {Wang}, \bibfnamefont {D.}}, \ and\ \bibinfo {author}
  {\bibnamefont {Yu}, \bibfnamefont {R.}},\ }\bibfield  {title} {\enquote
  {\bibinfo {title} {Automatic symmetry discovery with lie algebra
  convolutional network},}\ }\href@noop {} {\bibfield  {journal} {\bibinfo
  {journal} {Advances in Neural Information Processing Systems}\ }\textbf
  {\bibinfo {volume} {34}},\ \bibinfo {pages} {2503--2515} (\bibinfo {year}
  {2021})}\BibitemShut {NoStop}%
\bibitem [{\citenamefont {Dierkes}\ and\ \citenamefont
  {Fla{\ss}kamp}(2021)}]{dierkes2021learning}%
  \BibitemOpen
  \bibfield  {author} {\bibinfo {author} {\bibnamefont {Dierkes}, \bibfnamefont
  {E.}}\ and\ \bibinfo {author} {\bibnamefont {Fla{\ss}kamp}, \bibfnamefont
  {K.}},\ }\bibfield  {title} {\enquote {\bibinfo {title} {Learning
  {Hamiltonian} {Systems} considering {System} {Symmetries} in {Neural}
  {Networks}},}\ }\href {\doibase 10.1016/j.ifacol.2021.11.080} {\bibfield
  {journal} {\bibinfo  {journal} {IFAC-PapersOnLine}\ }\textbf {\bibinfo
  {volume} {54}},\ \bibinfo {pages} {210--216} (\bibinfo {year} {2021})},\
  \bibinfo {note} {7th IFAC Workshop on Lagrangian and Hamiltonian Methods for
  Nonlinear Control LHMNC 2021}\BibitemShut {NoStop}%
\bibitem [{\citenamefont {Finzi}\ \emph {et~al.}(2020)\citenamefont {Finzi},
  \citenamefont {Stanton}, \citenamefont {Izmailov},\ and\ \citenamefont
  {Wilson}}]{finzi2020generalizing}%
  \BibitemOpen
  \bibfield  {author} {\bibinfo {author} {\bibnamefont {Finzi}, \bibfnamefont
  {M.}}, \bibinfo {author} {\bibnamefont {Stanton}, \bibfnamefont {S.}},
  \bibinfo {author} {\bibnamefont {Izmailov}, \bibfnamefont {P.}}, \ and\
  \bibinfo {author} {\bibnamefont {Wilson}, \bibfnamefont {A.}},\ }\bibfield
  {title} {\enquote {\bibinfo {title} {Generalizing convolutional neural
  networks for equivariance to lie groups on arbitrary continuous data},}\ }in\
  \href@noop {} {\emph {\bibinfo {booktitle} {37th International Conference on
  Machine Learning, ICML 2020}}},\ \bibinfo {editor} {edited by\ \bibinfo
  {editor} {\bibfnamefont {H.}~\bibnamefont {Daume}}\ and\ \bibinfo {editor}
  {\bibfnamefont {A.}~\bibnamefont {Singh}}}\ (\bibinfo {year} {2020})\ pp.\
  \bibinfo {pages} {3146--3157}\BibitemShut {NoStop}%
\bibitem [{\citenamefont {Goodfellow}, \citenamefont {Bengio},\ and\
  \citenamefont {Courville}(2016)}]{Goodfellow-et-al-2016}%
  \BibitemOpen
  \bibfield  {author} {\bibinfo {author} {\bibnamefont {Goodfellow},
  \bibfnamefont {I.}}, \bibinfo {author} {\bibnamefont {Bengio}, \bibfnamefont
  {Y.}}, \ and\ \bibinfo {author} {\bibnamefont {Courville}, \bibfnamefont
  {A.}},\ }\href@noop {} {\emph {\bibinfo {title} {Deep Learning}}}\ (\bibinfo
  {publisher} {MIT Press},\ \bibinfo {year} {2016})\ \bibinfo {note}
  {\url{http://www.deeplearningbook.org}}\BibitemShut {NoStop}%
\bibitem [{\citenamefont {Greydanus}, \citenamefont {Dzamba},\ and\
  \citenamefont {Yosinski}(2019)}]{greydanus_hamiltonian_2019}%
  \BibitemOpen
  \bibfield  {author} {\bibinfo {author} {\bibnamefont {Greydanus},
  \bibfnamefont {S.}}, \bibinfo {author} {\bibnamefont {Dzamba}, \bibfnamefont
  {M.}}, \ and\ \bibinfo {author} {\bibnamefont {Yosinski}, \bibfnamefont
  {J.}},\ }\bibfield  {title} {\enquote {\bibinfo {title} {Hamiltonian {Neural}
  {Networks}},}\ }in\ \href
  {https://proceedings.neurips.cc/paper/2019/file/26cd8ecadce0d4efd6cc8a8725cbd1f8-Paper.pdf}
  {\emph {\bibinfo {booktitle} {Advances in {Neural} {Information} {Processing}
  {Systems}}}},\ Vol.~\bibinfo {volume} {32}\ (\bibinfo  {publisher} {Curran
  Associates, Inc.},\ \bibinfo {address} {Canada},\ \bibinfo {year} {2019})\
  pp.\ \bibinfo {pages} {15353--15363}\BibitemShut {NoStop}%
\bibitem [{\citenamefont {Griewank}\ and\ \citenamefont
  {Walther}(2008)}]{griewank2008evaluating}%
  \BibitemOpen
  \bibfield  {author} {\bibinfo {author} {\bibnamefont {Griewank},
  \bibfnamefont {A.}}\ and\ \bibinfo {author} {\bibnamefont {Walther},
  \bibfnamefont {A.}},\ }\href {\doibase 10.1137/1.9780898717761} {\emph
  {\bibinfo {title} {{Evaluating derivatives: principles and techniques of
  algorithmic differentiation}}}}\ (\bibinfo  {publisher} {Society for
  Industrial and Applied Mathematics},\ \bibinfo {year} {2008})\BibitemShut
  {NoStop}%
\bibitem [{\citenamefont {Haier}, \citenamefont {Lubich},\ and\ \citenamefont
  {Wanner}(2006)}]{hairer_geometric_2006}%
  \BibitemOpen
  \bibfield  {author} {\bibinfo {author} {\bibnamefont {Haier}, \bibfnamefont
  {E.}}, \bibinfo {author} {\bibnamefont {Lubich}, \bibfnamefont {C.}}, \ and\
  \bibinfo {author} {\bibnamefont {Wanner}, \bibfnamefont {G.}},\ }\href
  {\doibase 10.1007/3-540-30666-8} {\emph {\bibinfo {title} {Geometric
  Numerical integration: structure-preserving algorithms for ordinary
  differential equations}}}\ (\bibinfo  {publisher} {Springer Berlin},\
  \bibinfo {year} {2006})\BibitemShut {NoStop}%
\bibitem [{\citenamefont {Hintze}\ and\ \citenamefont
  {Nelson}(1998)}]{ViolinPlots}%
  \BibitemOpen
  \bibfield  {author} {\bibinfo {author} {\bibnamefont {Hintze}, \bibfnamefont
  {J.~L.}}\ and\ \bibinfo {author} {\bibnamefont {Nelson}, \bibfnamefont
  {R.~D.}},\ }\bibfield  {title} {\enquote {\bibinfo {title} {Violin plots: A
  box plot-density trace synergism},}\ }\href {\doibase
  10.1080/00031305.1998.10480559} {\bibfield  {journal} {\bibinfo  {journal}
  {The American Statistician}\ }\textbf {\bibinfo {volume} {52}},\ \bibinfo
  {pages} {181--184} (\bibinfo {year} {1998})},\ \Eprint
  {http://arxiv.org/abs/https://www.tandfonline.com/doi/pdf/10.1080/00031305.1998.10480559}
  {https://www.tandfonline.com/doi/pdf/10.1080/00031305.1998.10480559}
  \BibitemShut {NoStop}%
\bibitem [{\citenamefont {Hornik}, \citenamefont {Stinchcombe},\ and\
  \citenamefont {White}(1989)}]{hornik_multilayer_1989}%
  \BibitemOpen
  \bibfield  {author} {\bibinfo {author} {\bibnamefont {Hornik}, \bibfnamefont
  {K.}}, \bibinfo {author} {\bibnamefont {Stinchcombe}, \bibfnamefont {M.}}, \
  and\ \bibinfo {author} {\bibnamefont {White}, \bibfnamefont {H.}},\
  }\bibfield  {title} {\enquote {\bibinfo {title} {Multilayer feedforward
  networks are universal approximators},}\ }\href {\doibase
  https://doi.org/10.1016/0893-6080(89)90020-8} {\bibfield  {journal} {\bibinfo
   {journal} {Neural networks}\ }\textbf {\bibinfo {volume} {2}},\ \bibinfo
  {pages} {359--366} (\bibinfo {year} {1989})}\BibitemShut {NoStop}%
\bibitem [{\citenamefont {Jin}\ \emph {et~al.}(2020)\citenamefont {Jin},
  \citenamefont {Zhang}, \citenamefont {Zhu}, \citenamefont {Tang},\ and\
  \citenamefont {Karniadakis}}]{SympNets}%
  \BibitemOpen
  \bibfield  {author} {\bibinfo {author} {\bibnamefont {Jin}, \bibfnamefont
  {P.}}, \bibinfo {author} {\bibnamefont {Zhang}, \bibfnamefont {Z.}}, \bibinfo
  {author} {\bibnamefont {Zhu}, \bibfnamefont {A.}}, \bibinfo {author}
  {\bibnamefont {Tang}, \bibfnamefont {Y.}}, \ and\ \bibinfo {author}
  {\bibnamefont {Karniadakis}, \bibfnamefont {G.~E.}},\ }\bibfield  {title}
  {\enquote {\bibinfo {title} {Sympnets: Intrinsic structure-preserving
  symplectic networks for identifying {H}amiltonian systems},}\ }\href
  {\doibase 10.1016/j.neunet.2020.08.017} {\bibfield  {journal} {\bibinfo
  {journal} {Neural Networks}\ }\textbf {\bibinfo {volume} {132}},\ \bibinfo
  {pages} {166--179} (\bibinfo {year} {2020})}\BibitemShut {NoStop}%
\bibitem [{\citenamefont {Kingma}\ and\ \citenamefont
  {Ba}(2015)}]{kingma_adam_2017}%
  \BibitemOpen
  \bibfield  {author} {\bibinfo {author} {\bibnamefont {Kingma}, \bibfnamefont
  {D.~P.}}\ and\ \bibinfo {author} {\bibnamefont {Ba}, \bibfnamefont {J.}},\
  }\bibfield  {title} {\enquote {\bibinfo {title} {Adam: {A} method for
  stochastic optimization},}\ }in\ \href {http://arxiv.org/abs/1412.6980}
  {\emph {\bibinfo {booktitle} {3rd International Conference on Learning
  Representations, {ICLR} 2015, San Diego, CA, USA, May 7-9, 2015, Conference
  Track Proceedings}}}\ (\bibinfo {year} {2015})\BibitemShut {NoStop}%
\bibitem [{\citenamefont {Libermann}\ and\ \citenamefont
  {Marle}(1987)}]{Libermann1987}%
  \BibitemOpen
  \bibfield  {author} {\bibinfo {author} {\bibnamefont {Libermann},
  \bibfnamefont {P.}}\ and\ \bibinfo {author} {\bibnamefont {Marle},
  \bibfnamefont {C.-M.}},\ }\enquote {\bibinfo {title} {Symplectic manifolds
  and poisson manifolds},}\ in\ \href {\doibase 10.1007/978-94-009-3807-6_3}
  {\emph {\bibinfo {booktitle} {Symplectic Geometry and Analytical
  Mechanics}}}\ (\bibinfo  {publisher} {Springer Netherlands},\ \bibinfo
  {address} {Dordrecht},\ \bibinfo {year} {1987})\ pp.\ \bibinfo {pages}
  {89--184}\BibitemShut {NoStop}%
\bibitem [{\citenamefont {Lishkova}\ \emph {et~al.}(2022)\citenamefont
  {Lishkova}, \citenamefont {Scherer}, \citenamefont {Ridderbusch},
  \citenamefont {Jamnik}, \citenamefont {Li{\`o}}, \citenamefont
  {Ober-Bl{\"o}baum},\ and\ \citenamefont {Offen}}]{lishkova_discrete_2022}%
  \BibitemOpen
  \bibfield  {author} {\bibinfo {author} {\bibnamefont {Lishkova},
  \bibfnamefont {Y.}}, \bibinfo {author} {\bibnamefont {Scherer}, \bibfnamefont
  {P.}}, \bibinfo {author} {\bibnamefont {Ridderbusch}, \bibfnamefont {S.}},
  \bibinfo {author} {\bibnamefont {Jamnik}, \bibfnamefont {M.}}, \bibinfo
  {author} {\bibnamefont {Li{\`o}}, \bibfnamefont {P.}}, \bibinfo {author}
  {\bibnamefont {Ober-Bl{\"o}baum}, \bibfnamefont {S.}}, \ and\ \bibinfo
  {author} {\bibnamefont {Offen}, \bibfnamefont {C.}},\ }\bibfield  {title}
  {\enquote {\bibinfo {title} {Discrete lagrangian neural networks with
  automatic symmetry discovery},}\ }\href {\doibase 10.48550/arXiv.2211.10830}
  {\bibfield  {journal} {\bibinfo  {journal} {arXiv preprint}\ } (\bibinfo
  {year} {2022}),\ 10.48550/arXiv.2211.10830}\BibitemShut {NoStop}%
\bibitem [{\citenamefont {Marsden}\ and\ \citenamefont
  {Abraham}(1978)}]{Marsden78}%
  \BibitemOpen
  \bibfield  {author} {\bibinfo {author} {\bibnamefont {Marsden}, \bibfnamefont
  {J.~E.}}\ and\ \bibinfo {author} {\bibnamefont {Abraham}, \bibfnamefont
  {R.}},\ }\href {http://resolver.caltech.edu/CaltechBOOK:1987.001} {\emph
  {\bibinfo {title} {Foundations of Mechanics}}},\ \bibinfo {edition} {2nd}\
  ed.\ (\bibinfo  {publisher} {Addison-Wesley Publishing Co.},\ \bibinfo
  {address} {Redwood City, CA.},\ \bibinfo {year} {1978})\BibitemShut {NoStop}%
\bibitem [{\citenamefont {Marsden}\ and\ \citenamefont
  {Ratiu}(1999)}]{MarsdenRatiu}%
  \BibitemOpen
  \bibfield  {author} {\bibinfo {author} {\bibnamefont {Marsden}, \bibfnamefont
  {J.~E.}}\ and\ \bibinfo {author} {\bibnamefont {Ratiu}, \bibfnamefont
  {T.~S.}},\ }\href {\doibase 10.1007/978-0-387-21792-5} {\emph {\bibinfo
  {title} {Introduction to Mechanics and Symmetry: A Basic Exposition of
  Classical Mechanical Systems}}}\ (\bibinfo  {publisher} {Springer New York},\
  \bibinfo {address} {New York, NY},\ \bibinfo {year} {1999})\BibitemShut
  {NoStop}%
\bibitem [{\citenamefont {Marsden}\ and\ \citenamefont
  {West}(2001)}]{marsden2001discrete}%
  \BibitemOpen
  \bibfield  {author} {\bibinfo {author} {\bibnamefont {Marsden}, \bibfnamefont
  {J.~E.}}\ and\ \bibinfo {author} {\bibnamefont {West}, \bibfnamefont {M.}},\
  }\bibfield  {title} {\enquote {\bibinfo {title} {Discrete mechanics and
  variational integrators},}\ }\href@noop {} {\bibfield  {journal} {\bibinfo
  {journal} {Acta Numerica}\ }\textbf {\bibinfo {volume} {10}},\ \bibinfo
  {pages} {357--514} (\bibinfo {year} {2001})}\BibitemShut {NoStop}%
\bibitem [{\citenamefont {Mason}\ \emph {et~al.}(2022)\citenamefont {Mason},
  \citenamefont {Allen-Blanchette}, \citenamefont {Zolman}, \citenamefont
  {Davison},\ and\ \citenamefont {Leonard}}]{mason2022learning}%
  \BibitemOpen
  \bibfield  {author} {\bibinfo {author} {\bibnamefont {Mason}, \bibfnamefont
  {J.}}, \bibinfo {author} {\bibnamefont {Allen-Blanchette}, \bibfnamefont
  {C.}}, \bibinfo {author} {\bibnamefont {Zolman}, \bibfnamefont {N.}},
  \bibinfo {author} {\bibnamefont {Davison}, \bibfnamefont {E.}}, \ and\
  \bibinfo {author} {\bibnamefont {Leonard}, \bibfnamefont {N.}},\ }\href@noop
  {} {\enquote {\bibinfo {title} {Learning interpretable dynamics from images
  of a freely rotating 3d rigid body},}\ } (\bibinfo {year} {2022}),\ \Eprint
  {http://arxiv.org/abs/2209.11355} {arXiv:2209.11355 [cs.CV]} \BibitemShut
  {NoStop}%
\bibitem [{\citenamefont {Ober-Blöbaum}\ and\ \citenamefont
  {Offen}(2022)}]{LSI}%
  \BibitemOpen
  \bibfield  {author} {\bibinfo {author} {\bibnamefont {Ober-Blöbaum},
  \bibfnamefont {S.}}\ and\ \bibinfo {author} {\bibnamefont {Offen},
  \bibfnamefont {C.}},\ }\bibfield  {title} {\enquote {\bibinfo {title}
  {Variational integration of learned dynamical systems},}\ }\href {\doibase
  10.1016/j.cam.2022.114780} {\bibfield  {journal} {\bibinfo  {journal}
  {Journal of Computational and Applied Mathematics}\ }\textbf {\bibinfo
  {volume} {421}} (\bibinfo {year} {2022}),\
  10.1016/j.cam.2022.114780}\BibitemShut {NoStop}%
\bibitem [{\citenamefont {Offen}\ and\ \citenamefont
  {Ober-Blöbaum}(2022)}]{SSI}%
  \BibitemOpen
  \bibfield  {author} {\bibinfo {author} {\bibnamefont {Offen}, \bibfnamefont
  {C.}}\ and\ \bibinfo {author} {\bibnamefont {Ober-Blöbaum}, \bibfnamefont
  {S.}},\ }\bibfield  {title} {\enquote {\bibinfo {title} {Symplectic
  integration of learned {H}amiltonian systems},}\ }\href {\doibase
  10.1063/5.0065913} {\bibfield  {journal} {\bibinfo  {journal} {Chaos: An
  Interdisciplinary Journal of Nonlinear Science}\ }\textbf {\bibinfo {volume}
  {32}} (\bibinfo {year} {2022}),\ 10.1063/5.0065913}\BibitemShut {NoStop}%
\bibitem [{\citenamefont {Offen}\ and\ \citenamefont
  {Ober-Blöbaum}(2023)}]{LDensityLearning}%
  \BibitemOpen
  \bibfield  {author} {\bibinfo {author} {\bibnamefont {Offen}, \bibfnamefont
  {C.}}\ and\ \bibinfo {author} {\bibnamefont {Ober-Blöbaum}, \bibfnamefont
  {S.}},\ }\href@noop {} {\enquote {\bibinfo {title} {Learning discrete
  {L}agrangians for variational {PDE}s from data and detection of travelling
  waves},}\ } (\bibinfo {year} {2023}),\ \Eprint
  {http://arxiv.org/abs/2302.08232} {arXiv:2302.08232 [math.NA]} \BibitemShut
  {NoStop}%
\bibitem [{\citenamefont {Olver}(1986)}]{Olver1986}%
  \BibitemOpen
  \bibfield  {author} {\bibinfo {author} {\bibnamefont {Olver}, \bibfnamefont
  {P.~J.}},\ }\href {\doibase 10.1007/978-1-4684-0274-2} {\emph {\bibinfo
  {title} {Applications of Lie Groups to Differential Equations}}}\ (\bibinfo
  {publisher} {Springer New York},\ \bibinfo {year} {1986})\ pp.\ \bibinfo
  {pages} {1--76}\BibitemShut {NoStop}%
\bibitem [{\citenamefont {Paszke}\ \emph {et~al.}(2019)\citenamefont {Paszke},
  \citenamefont {Gross}, \citenamefont {Massa}, \citenamefont {Lerer},
  \citenamefont {Bradbury}, \citenamefont {Chanan}, \citenamefont {Killeen},
  \citenamefont {Lin}, \citenamefont {Gimelshein}, \citenamefont {Antiga} \emph
  {et~al.}}]{paszke2019pytorch}%
  \BibitemOpen
  \bibfield  {author} {\bibinfo {author} {\bibnamefont {Paszke}, \bibfnamefont
  {A.}}, \bibinfo {author} {\bibnamefont {Gross}, \bibfnamefont {S.}}, \bibinfo
  {author} {\bibnamefont {Massa}, \bibfnamefont {F.}}, \bibinfo {author}
  {\bibnamefont {Lerer}, \bibfnamefont {A.}}, \bibinfo {author} {\bibnamefont
  {Bradbury}, \bibfnamefont {J.}}, \bibinfo {author} {\bibnamefont {Chanan},
  \bibfnamefont {G.}}, \bibinfo {author} {\bibnamefont {Killeen}, \bibfnamefont
  {T.}}, \bibinfo {author} {\bibnamefont {Lin}, \bibfnamefont {Z.}}, \bibinfo
  {author} {\bibnamefont {Gimelshein}, \bibfnamefont {N.}}, \bibinfo {author}
  {\bibnamefont {Antiga}, \bibfnamefont {L.}},  \emph {et~al.},\ }\bibfield
  {title} {\enquote {\bibinfo {title} {Pytorch: An imperative style,
  high-performance deep learning library},}\ }in\ \href
  {https://dl.acm.org/doi/10.5555/3454287.3455008} {\emph {\bibinfo {booktitle}
  {Proceedings of the 33rd International Conference on Neural Information
  Processing Systems}}},\ Vol.~\bibinfo {volume} {32}\ (\bibinfo  {publisher}
  {Curran Associates Inc.},\ \bibinfo {year} {2019})\BibitemShut {NoStop}%
\bibitem [{\citenamefont {Pedrosa}, \citenamefont {Schneider},\ and\
  \citenamefont {Flaßkamp}(2022)}]{pedrosa22learning}%
  \BibitemOpen
  \bibfield  {author} {\bibinfo {author} {\bibnamefont {Pedrosa}, \bibfnamefont
  {M.~V.~A.}}, \bibinfo {author} {\bibnamefont {Schneider}, \bibfnamefont
  {T.}}, \ and\ \bibinfo {author} {\bibnamefont {Flaßkamp}, \bibfnamefont
  {K.}},\ }\bibfield  {title} {\enquote {\bibinfo {title} {Learning motion
  primitives automata for autonomous driving applications},}\ }\href {\doibase
  10.3390/mca27040054} {\bibfield  {journal} {\bibinfo  {journal} {Mathematical
  and Computational Applications}\ }\textbf {\bibinfo {volume} {27}} (\bibinfo
  {year} {2022}),\ 10.3390/mca27040054}\BibitemShut {NoStop}%
\bibitem [{\citenamefont {Raissi}, \citenamefont {Perdikaris},\ and\
  \citenamefont {Karniadakis}(2019)}]{raissi_physics-informed_2019}%
  \BibitemOpen
  \bibfield  {author} {\bibinfo {author} {\bibnamefont {Raissi}, \bibfnamefont
  {M.}}, \bibinfo {author} {\bibnamefont {Perdikaris}, \bibfnamefont {P.}}, \
  and\ \bibinfo {author} {\bibnamefont {Karniadakis}, \bibfnamefont {G.~E.}},\
  }\bibfield  {title} {\enquote {\bibinfo {title} {Physics-informed neural
  networks: {A} deep learning framework for solving forward and inverse
  problems involving nonlinear partial differential equations},}\ }\href
  {\doibase 10.1016/j.jcp.2018.10.045} {\bibfield  {journal} {\bibinfo
  {journal} {Journal of Computational Physics}\ }\textbf {\bibinfo {volume}
  {378}},\ \bibinfo {pages} {686--707} (\bibinfo {year} {2019})}\BibitemShut
  {NoStop}%
\bibitem [{\citenamefont {Rath}\ \emph {et~al.}(2021)\citenamefont {Rath},
  \citenamefont {Albert}, \citenamefont {Bischl},\ and\ \citenamefont {von
  Toussaint}}]{Rath2021}%
  \BibitemOpen
  \bibfield  {author} {\bibinfo {author} {\bibnamefont {Rath}, \bibfnamefont
  {K.}}, \bibinfo {author} {\bibnamefont {Albert}, \bibfnamefont {C.~G.}},
  \bibinfo {author} {\bibnamefont {Bischl}, \bibfnamefont {B.}}, \ and\
  \bibinfo {author} {\bibnamefont {von Toussaint}, \bibfnamefont {U.}},\
  }\bibfield  {title} {\enquote {\bibinfo {title} {{S}ymplectic {G}aussian
  process regression of maps in {H}amiltonian systems},}\ }\href {\doibase
  10.1063/5.0048129} {\bibfield  {journal} {\bibinfo  {journal} {Chaos: An
  Interdisciplinary Journal of Nonlinear Science}\ }\textbf {\bibinfo {volume}
  {31}} (\bibinfo {year} {2021}),\ 10.1063/5.0048129}\BibitemShut {NoStop}%
\bibitem [{\citenamefont {Ridderbusch}\ \emph {et~al.}(2021)\citenamefont
  {Ridderbusch}, \citenamefont {Offen}, \citenamefont {Ober-Blöbaum},\ and\
  \citenamefont {Goulart}}]{Ridderbusch2021}%
  \BibitemOpen
  \bibfield  {author} {\bibinfo {author} {\bibnamefont {Ridderbusch},
  \bibfnamefont {S.}}, \bibinfo {author} {\bibnamefont {Offen}, \bibfnamefont
  {C.}}, \bibinfo {author} {\bibnamefont {Ober-Blöbaum}, \bibfnamefont {S.}},
  \ and\ \bibinfo {author} {\bibnamefont {Goulart}, \bibfnamefont {P.}},\
  }\bibfield  {title} {\enquote {\bibinfo {title} {Learning {ODE} models with
  qualitative structure using {G}aussian processes},}\ }\href {\doibase
  10.1109/cdc45484.2021.9683426} {\bibfield  {journal} {\bibinfo  {journal}
  {2021 60th IEEE Conference on Decision and Control (CDC)}\ } (\bibinfo {year}
  {2021}),\ 10.1109/cdc45484.2021.9683426}\BibitemShut {NoStop}%
\bibitem [{\citenamefont {Runge}(1895)}]{runge_uber_1895}%
  \BibitemOpen
  \bibfield  {author} {\bibinfo {author} {\bibnamefont {Runge}, \bibfnamefont
  {C.}},\ }\bibfield  {title} {\enquote {\bibinfo {title} {Über die numerische
  {Auflösung} von {Differentialgleichungen}},}\ }\href {\doibase
  10.1007/BF01446807} {\bibfield  {journal} {\bibinfo  {journal} {Mathematische
  Annalen}\ }\textbf {\bibinfo {volume} {46}},\ \bibinfo {pages} {167--178}
  (\bibinfo {year} {1895})}\BibitemShut {NoStop}%
\bibitem [{\citenamefont {So}\ \emph {et~al.}(2022)\citenamefont {So},
  \citenamefont {Li}, \citenamefont {Theodorou},\ and\ \citenamefont
  {Tao}}]{So22datadriven}%
  \BibitemOpen
  \bibfield  {author} {\bibinfo {author} {\bibnamefont {So}, \bibfnamefont
  {O.}}, \bibinfo {author} {\bibnamefont {Li}, \bibfnamefont {G.}}, \bibinfo
  {author} {\bibnamefont {Theodorou}, \bibfnamefont {E.~A.}}, \ and\ \bibinfo
  {author} {\bibnamefont {Tao}, \bibfnamefont {M.}},\ }\bibfield  {title}
  {\enquote {\bibinfo {title} {Data-driven discovery of non-newtonian astronomy
  via learning non-euclidean hamiltonian},}\ }\href {\doibase
  10.48550/arXiv.2210.00090} {\bibfield  {journal} {\bibinfo  {journal} {arXiv
  preprint}\ } (\bibinfo {year} {2022}),\
  10.48550/arXiv.2210.00090}\BibitemShut {NoStop}%
\bibitem [{\citenamefont {Udrescu}\ and\ \citenamefont
  {Tegmark}(2020)}]{udrescu_ai_2020}%
  \BibitemOpen
  \bibfield  {author} {\bibinfo {author} {\bibnamefont {Udrescu}, \bibfnamefont
  {S.-M.}}\ and\ \bibinfo {author} {\bibnamefont {Tegmark}, \bibfnamefont
  {M.}},\ }\bibfield  {title} {\enquote {\bibinfo {title} {{AI Feynman: A
  physics-inspired method for symbolic regression}},}\ }\href {\doibase
  10.1126/sciadv.aay2631} {\bibfield  {journal} {\bibinfo  {journal} {Science
  Advances}\ }\textbf {\bibinfo {volume} {6}} (\bibinfo {year} {2020}),\
  10.1126/sciadv.aay2631}\BibitemShut {NoStop}%
\bibitem [{\citenamefont {Vallado}(2001)}]{vallado2001fundamentals}%
  \BibitemOpen
  \bibfield  {author} {\bibinfo {author} {\bibnamefont {Vallado}, \bibfnamefont
  {D.~A.}},\ }\href@noop {} {\emph {\bibinfo {title} {Fundamentals of
  astrodynamics and applications}}},\ Vol.~\bibinfo {volume} {12}\ (\bibinfo
  {publisher} {Springer Science \& Business Media},\ \bibinfo {year}
  {2001})\BibitemShut {NoStop}%
\bibitem [{\citenamefont {Zhong}, \citenamefont {Dey},\ and\ \citenamefont
  {Chakraborty}(2020)}]{Zhong2020Symplectic}%
  \BibitemOpen
  \bibfield  {author} {\bibinfo {author} {\bibnamefont {Zhong}, \bibfnamefont
  {Y.~D.}}, \bibinfo {author} {\bibnamefont {Dey}, \bibfnamefont {B.}}, \ and\
  \bibinfo {author} {\bibnamefont {Chakraborty}, \bibfnamefont {A.}},\
  }\bibfield  {title} {\enquote {\bibinfo {title} {Symplectic ode-net: Learning
  hamiltonian dynamics with control},}\ }in\ \href
  {https://openreview.net/forum?id=ryxmb1rKDS} {\emph {\bibinfo {booktitle}
  {International Conference on Learning Representations}}}\ (\bibinfo {year}
  {2020})\BibitemShut {NoStop}%
\end{thebibliography}%

\end{document}